%% file: arxiv.tex
\documentclass[10pt,twocolumn,letterpaper]{article}

\usepackage[pagenumbers]{iccv} %

\input{preamble}

\definecolor{iccvblue}{rgb}{0.21,0.49,0.74}
\usepackage[pagebackref,breaklinks,colorlinks,allcolors=iccvblue]{hyperref}

\newcommand{\foobar}[1]{}

\def\paperID{*****} %
\def\confName{ICCV}
\def\confYear{2025}

\title{Unleashing Vecset Diffusion Model for Fast Shape Generation}

\author{
    Zeqiang Lai$^{1,2\star}$
    , Yunfei Zhao$^{2,3\star}$ 
    , Zibo Zhao$^{2,4}$ 
    , Haolin Liu$^{2}$ 
    , Fuyun Wang$^{1}$ \\
     Huiwen Shi$^{2}$ 
    , Xianghui Yang$^{2}$ 
    , Qingxiang Lin$^{2}$ 
    , Jingwei Huang$^{2}$ \\
      Yuhong Liu$^{2}$
    , Jie Jiang$^{2}$
    , Chunchao Guo$^{2\dagger}$ 
    , Xiangyu Yue$^{1\dagger}$ \\
	$^1${MMLab, CUHK} \quad $^2${Tencent Hunyuan} \quad $^3$VISG, NJU \quad $^4${ShanghaiTech}\\
    \url{https://github.com/Tencent/FlashVDM}
}

\begin{document}

\twocolumn[{%
\renewcommand\twocolumn[1][]{#1}%
\maketitle
\begin{center}
    \centering
    \captionsetup{type=figure}
    \includegraphics[width=\linewidth]{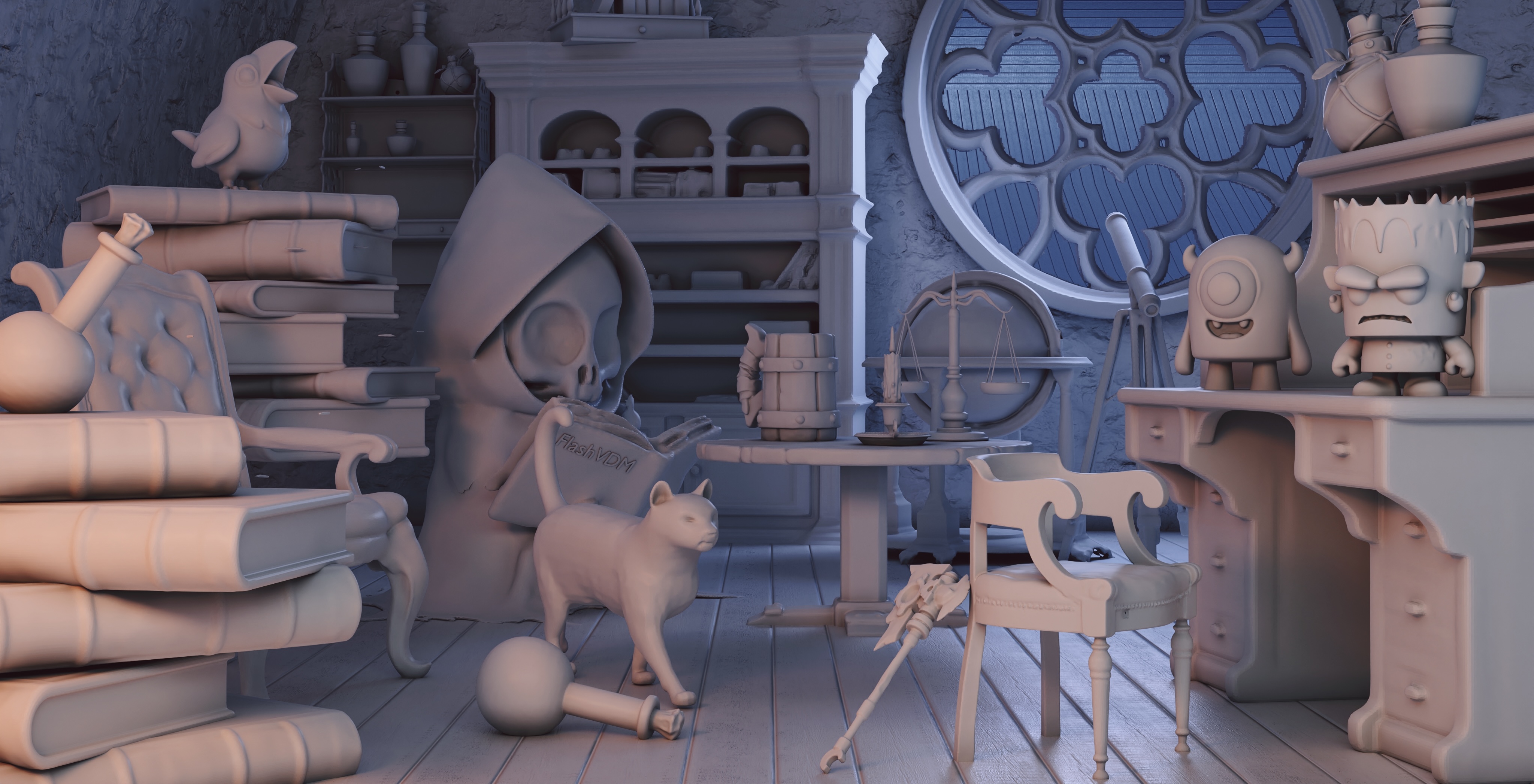}
    \captionof{figure}{
    High-resolution 3D shapes generated by Flash Vecset Diffusion Model (\shortname) within 1 second. 
    }
    \label{fig:teaser}
\end{center}%
}]

\def\thefootnote{}\footnotetext{$\star$ Equal contribution. $\dagger$ Corresponding authors.} 

\input{sec/0_abstract}    
\input{sec/1_intro}
\input{sec/2_related_work}

\input{sec/4_method}

\input{sec/5_experiment}
\input{sec/6_conclusion}

\clearpage
\appendix
\input{sec/x_appendix}
{
    \small
    \bibliographystyle{ieeenat_fullname}
    \bibliography{main}
}

\end{document}

%% file: preamble.tex
\usepackage{csquotes}
\usepackage{blindtext}
\usepackage{algorithm}
\usepackage{algpseudocode}
\usepackage{amsmath}
\usepackage{xspace}
\usepackage{tcolorbox}
\usepackage{lipsum}
\algrenewcommand\algorithmicrequire{\textbf{Input:}}
\algrenewcommand\algorithmicensure{\textbf{Output:}}

\usepackage{makecell}
\newcommand{\shortname}{FlashVDM\xspace}
\newcommand{\vdname}{Hierarchical Volume Decoding\xspace}
\newcommand{\lvdname}{hierarchical volume decoding\xspace}
\newcommand{\svdname}{hierarchical decoding\xspace}
\newcommand{\scvdname}{Hierarchical Decoding\xspace}
\newcommand{\hyshape}{Hunyuan3D-2\xspace}
\newcommand{\vaespeedup}{45\texttimes\xspace}
\newcommand{\ditspeedup}{32\texttimes\xspace}

\newcommand{\queryreduce}{91.4\%\xspace}

\newcommand{\caflopsreduce}{97.1\%\xspace}

\usepackage{colortbl}
\definecolor{graycolor}{rgb}{0.95,0.95,0.95}

%% file: sec/0_abstract.tex
\begin{abstract}

3D shape generation has greatly flourished through the development of so-called ``native" 3D diffusion, particularly through the Vecset Diffusion Model (VDM). While recent advancements have shown promising results in generating high-resolution 3D shapes, VDM still struggles at high-speed generation. 
Challenges exist because of not only difficulties in accelerating diffusion sampling but also VAE decoding in VDM -- areas under-explored in previous works.
To address these challenges, we present \textbf{\shortname}, a systematic framework for accelerating both VAE and DiT in VDM.
For DiT, \shortname enables flexible diffusion sampling with as few as 5 inference steps and comparable quality, which is made possible by stabilizing consistency distillation with our newly introduced Progressive Flow Distillation.
For VAE, we introduce a lightning vecset decoder equipped with Adaptive KV Selection, \vdname, and Efficient Network Design. By exploiting the locality of vecset and the sparsity of shape surface in the volume, our decoder drastically lowers FLOPs, minimizing the overall decoding overhead.
We apply \shortname to \hyshape~\cite{zhao2025hunyuan3d} to obtain \hyshape Turbo.  Through systematic evaluation, we show that our model significantly outperforms existing fast 3D generation methods, achieving comparable performance to the state-of-the-art while reducing inference time by over \textbf{\vaespeedup} for reconstruction and \textbf{\ditspeedup} for generation. 
\end{abstract} 

%% file: sec/1_intro.tex
\section{Introduction}

3D shape generation has greatly flourished through the development of so-called ``native” 3D diffusion~\cite{zhang2024clay,xiang2024structured}, among which the Vecset Diffusion Model (VDM)~\cite{zhang20233dshape2vecset,zhao2024michelangelo} receives prominent attention and applications due to its simplicity and scalability. While recent works~\cite{zhang2024clay,zhao2025hunyuan3d,li2025triposg} have demonstrated the capabilities of VDM in generating high-resolution and high-quality 3D shapes, VDM still struggles with high-speed generation, typically requiring over 30 seconds$^{1}$ per shape\footnote{$^{1}$Measured by the default setting of Hunyuan3D-2~\cite{zhao2025hunyuan3d}.} -- far behind the development of image generation counterparts~\cite{yin2024onestep,sauer2023adversarial,luo2023latent,sauer2024fast}.
The notable reasons behind the lags of VDM~\cite{zhang20233dshape2vecset} against previous 2D predecessors~\cite{rombach2022high} lies in not only the lack of research of diffusion distillation~\cite{meng2023distillation,luo2023latent} for few-step 3D generators, but also the acceleration of Variational Autoencoders (VAE). 

\begin{figure}[t]
  \centering
  \includegraphics[width=\linewidth]{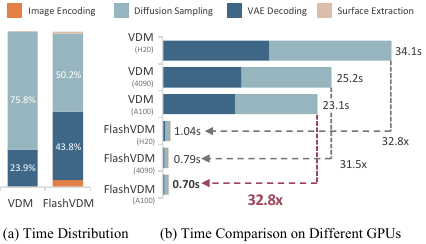}
   \caption{(a) VDM exhibits different  distribution with image DM with much larger percentage in VAE decoding. (b) FlashVDM enables fast shape generation within 1 second on a consumer GPU.}
   \label{fig:time_dist}
\end{figure}

Unlike the 2D VAE~\cite{esser2021taming}, which utilizes convolution for image compressing and decoding with structured latent, the VAE in VDM takes a fundamentally different approach. 
Its encoder compresses the point cloud of the mesh surface into a set of latent tokens---typically referred to as a \textit{vecset}---through cross-attention~(CA), using a set of query tokens~\cite{zhang20233dshape2vecset}. This approach, similar to Q-former~\cite{li2023blip} and Perceiver~\cite{jaegle2021perceiver}, offers strong compression capability but also introduces challenges in decoding.
To overcome this, VDM employs a symmetric approach, constructing a set of volume queries that evaluate occupancy or Signed Distance Function (SDF) at each grid point using another CA.
While the CA decoder design enables VDM to flexibly decode at arbitrary resolutions, the computational cost increases cubically with respect to resolution. 
For instance, at a typical resolution of 384, the number of query points exceeds 56 million---each requires a CA operation. Even worse, modern VDMs~\cite{zhao2025hunyuan3d, li2025triposg} typically use a much larger set of latent tokens for key-value pairs to ensure high-resolution generation, which ends up making the decoding process even slower.
As a result, VAE decoding in VDM demands a significant amount of inference time---even longer than diffusion sampling as shown in \cref{fig:time_dist} (a).

On the other hand, despite the significant advancements in fast image generation through diffusion distillation~\cite{meng2023distillation, yin2024onestep,sauer2023adversarial,luo2023latent,sauer2024fast}, pushing the speed to real-time, 
research into native 3D diffusion distillation remains scarce. 
Most existing distillation methods~\cite{yin2024onestep,sauer2023adversarial,luo2023latent} were originally designed for images, with some adaptations made for video~\cite{zhu2024accelerating,zhai2025motion}. 
Many techniques, such as LPIPS loss in Consistency Distillation~(CD)~\cite{song2023consistency,song2023improved,lu2024simplifying} and GAN designs~\cite{wang2025phased,yin2024onestep}, are specifically tailored for images.
However, the representation of 3D meshes and images differs significantly, making it challenging to adapt these techniques. Additionally, the vastly different latent spaces in VDMs against image DMs~\cite{rombach2022high} could drastically alter the training dynamics of the distillation, which makes previous strategies unsuitable, potentially leading to instability and unsatisfactory results.

To address the aforementioned problems, we present \textit{\shortname}, a framework for transforming pretrained VDM into a high-fidelity and high-speed generator with over \textbf{\vaespeedup speedup} in VAE decoding and \textbf{\ditspeedup speedup} overall. 
Our framework consists of two main components: 
1) diffusion acceleration and 2) VAE acceleration. 

For diffusion acceleration, we carefully analyzed the problems in direct application of CD~\cite{song2023consistency,wang2025phased} for VDM and found that the core issue lies in the instability of the target network, which degrades the training and causes unsatisfactory results. As a remedy, we propose \textit{Progressive Flow Distillation}, a multi-stage distillation method for VDM. It first decouples the CD distillation by warming up with guidance distillation. This prevents the target bias and fluctuations from misleading the student. 
Then, we carefully ablate the design choices including loss and training strategies in CD, leading to even more stable step distillation.
Finally, we introduce adversarial finetuning, leveraging real 3D data as supervision, to help distilled models produce smoother and more accurate results within limited steps. 
Together, our distilled model could achieve comparable results with only 5 Number of Function Evaluation (NFE).

For VAE acceleration, we introduce three techniques, including two training-free methods---\textit{Adaptive KV Selection} and \textit{\vdname}---to reduce both the number of number of key-value pairs and the queries. 
These techniques effectively exploit the locality of VDM point-queries and the sparsity of shape surface in the volume space, resulting in a highly efficient CA with a \caflopsreduce FLOPs reduction in total. 
Additionally, we introduce an \textit{Efficient Decoder Architecture}. By fixing the encoder and fine-tuning only the decoder, our new design further reduces FLOPs by 76.6\% for each query.
With these three improvements combined, our final VAE decoder achieves over a \vaespeedup speedup, significantly reducing the decoding time from 22.3s to just 0.49s.

In summary, our contributions are listed as follows:
\begin{itemize}[leftmargin=*]
    \item We introduce \shortname, a framework that converts pretrained VDMs into high-speed and -fidelity generators. 
    \item We present progressive flow distillation for VDM, significantly improving the stability and quality of the distilled model, achieving comparable results with only 5 NFE.
    \item We propose a novel algorithm, \lvdname with adaptive KV selection and an efficient decoder architecture, obtaining \vaespeedup speedup in decoding.
    \item We apply \shortname to \hyshape~\cite{zhao2025hunyuan3d} and obtain \hyshape Turbo, a high-resolution shape generator that matches its teacher’s shape quality with over \ditspeedup speedup, achieving 1 second per shape.
\end{itemize}

%% file: sec/2_related_work.tex
\section{Related Works}

\textbf{Diffusion Acceleration.} 
Step distillation~\cite{salimans2022progressive, meng2023distillation, yan2024perflow, luhman2021knowledge, song2023consistency, sauer2023adversarial, yin2024onestep, heek2024multistep, xu2024accelerating, zhou2024score} is a popular technique for accelerating diffusion models, with early works like Progressive Distillation~\cite{salimans2022progressive, meng2023distillation} training student models by halving sampling steps step-by-step. More recent approaches, such as consistency models~\cite{song2023consistency, wang2025phased, song2024improved} and score distillation~\cite{yin2024onestep, sauer2023adversarial}, enforce self-consistency or match distributions between student and teacher models. GAN-based approaches~\cite{sauer2023adversarial, wang2022diffusion} further refine score distillation by augmenting loss functions with adversarial objectives. Several works, including UFOGen~\cite{xu2023ufogen} and LADD~\cite{sauer2024fast}, combine GANs with pretrained diffusion models as discriminators. Other research focuses on reducing redundancy in sampling processes, such as DPM-Solver~\cite{lu2022dpm},  SnapFusion~\cite{li2023snapfusion}, and DeepCache~\cite{ma2024deepcache}. In this work, we introduce progressive flow distillation to address practical issues in 3D foundation models, an area not yet explored in previous research.

\textbf{VAE Acceleration.} 
The proposed \lvdname is generally related to octree decoding~\cite{saito2019pifu}. However, we noted that vanilla octree decoding introduces artifacts and holes, whereas our method is nearly lossless. Our adaptive KV selection is broadly related to previous work on token merging~\cite{bolya2023token, bolya2022token} in DiT inference, though our approach is specifically designed for VAE acceleration in VDM without the merge operation. To the best of our knowledge, we are the first to explore the efficient design of a VDM decoder. DC-AE~\cite{chen2024deep} is one of the works for image diffusion models~\cite{chen2024deep}, but it focuses on achieving a higher compression ratio, which is different from our objective.

\textbf{Fast 3D Generation.} 
Previous works on fast 3D generation are based on feedforward 3D reconstruction methods~\cite{hong2023lrm}, which require only a single network evaluation. Benefiting from this property, TripoSR~\cite{tochilkin2024triposr} and SF3D~\cite{boss2024sf3d} can generate a mesh from a single image in under 1 second, but the resulting mesh quality is limited. To address this, multiview images are often used~\cite{tang2024lgm,li2023instant3d}, but this requires a Multi-View Diffusion (MVD) model~\cite{shi2023zero123pp,shi2023mvdream,liu2023zero}. In response, SPAR3D~\cite{huang2025spar3d} replaces MVD with a small point cloud generator, while Turbo3D~\cite{hu2024turbo3d} distills MVD into a four-step generator. Nevertheless, all these feedforward methods are still limited in mesh quality compared to diffusion-based 3D generators~\cite{zhao2025hunyuan3d, zhang2024clay}.

%% file: sec/4_method.tex
\section{Method}

In this section, we first provide an introduction to VDM pipeline and inference time distribution. Then, we illustrate our approaches for accelerating VAE decoding in \cref{sec:lvd} and diffusion sampling in \cref{sec:pfd}, respectively.

\begin{figure}[t]
  \centering
  \includegraphics[width=\linewidth]{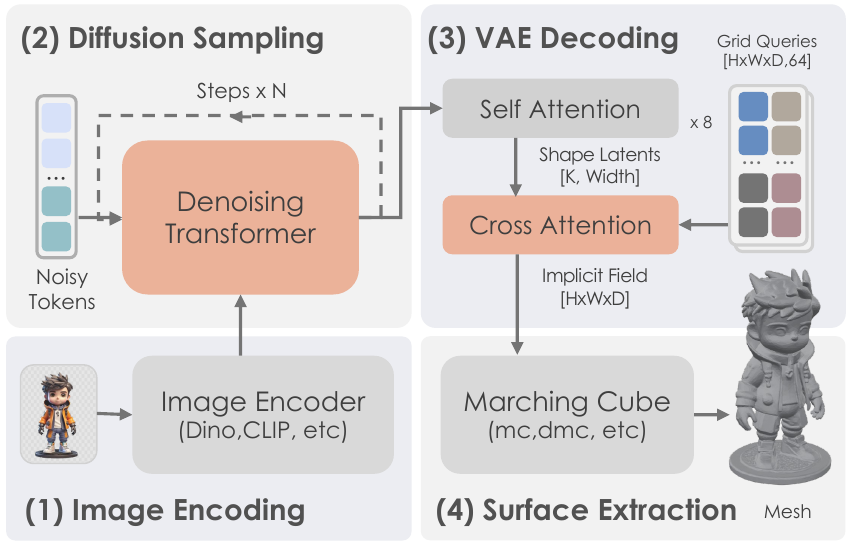}
   \caption{Illustration of four main stages of VDM. }
   \label{fig:vdm}
\end{figure}

\subsection{Preliminary of VDM}
\textbf{Pipeline Overview.} Here, we provide a brief introduction to the inference pipeline of VDM. We refer interested readers to~\cite{zhang20233dshape2vecset,zhang2024clay,chen2024dora,zhao2025hunyuan3d} for more details. As shown in \cref{fig:vdm}, the inference of VDM consists of four stages: 1) image encoding: an image encoder~\cite{oquab2023dinov2,radford2021learning} is used to extract conditional features; 2) diffusion sampling: iteratively calling a denoising transformer to predict shape latent based on conditional features. 3) volume decoding: the predicted shape latent is decoded into volume SDF via several layers of self-attention and a cross-attention. 4) surface extraction: marching cube algorithm~\cite{lorensen1998marching} is used to extract polygon mesh from the decoded volume.

\textbf{Inference Time Distribution.}
The distribution of inference times is shown in \cref{fig:time_dist}(a). We could observe that diffusion sampling (23.9\%) and volume decoding (75.8\%) account for the majority (99.7\%) of the inference time for VDM. However, image encoding and surface extraction remain significant when the total inference time is reduced to less than one second.
As a first step, we implement several engineering acceleration techniques, including FP8 attention with SageAttention2~\cite{zhang2024sageattention2}, static graph optimization with \texttt{torch.compile}, and \etc, which decrease the processing time to 25.09s per shape, as shown in \cref{fig:step-by-step-time}.

\subsection{Lightning Vecset Decoder}
\label{sec:lvd}

The majority (99.9\%) of the computational cost in the original VDM decoder is concentrated in its final cross-attention layer, which is evaluated tens of millions of times. To optimize this layer, we propose three key techniques, each addressing a different aspect: 1) \lvdname to reduce the number of queries, 2) adaptive KV selection to minimize the number of keys and values, and 3) an efficient decoder design to reduce computational overhead in each cross-attention operation.

\begin{figure}[t]
  \centering
  \includegraphics[width=\linewidth]{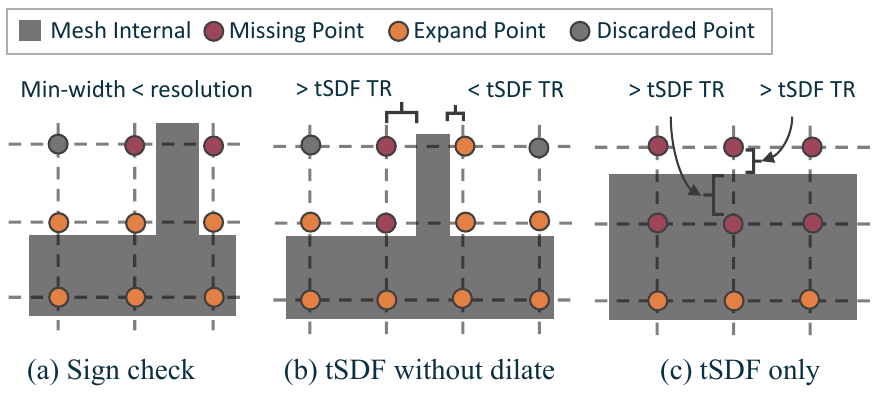}
  \foobar{-6mm}
   \caption{The corner cases in \lvdname. }
   \foobar{-2mm}
   \label{fig:ovd_corner_cases}
\end{figure}

\textbf{\vdname.} 
To decode shape latents into a mesh, the existing VDM decoder relies on dense volume decoding and marching cube algorithms~\cite{lorensen1998marching}, which has cubic complexity with respect to the resolution. It raises challenges for balancing speed and mesh quality. However, unlike image decoders~\cite{esser2021taming} that must predict RGB values for every pixel, the VDM decoder only needs to determine high-resolution SDF values near the shape surface -- all voxels far from the surface can be simply classified as inside or outside. Inspired by this characteristic, we propose \lvdname that only increases resolution near the surface, thereby reducing the queries by \queryreduce.

\begin{figure}[t]
  \centering
  \includegraphics[width=\linewidth]{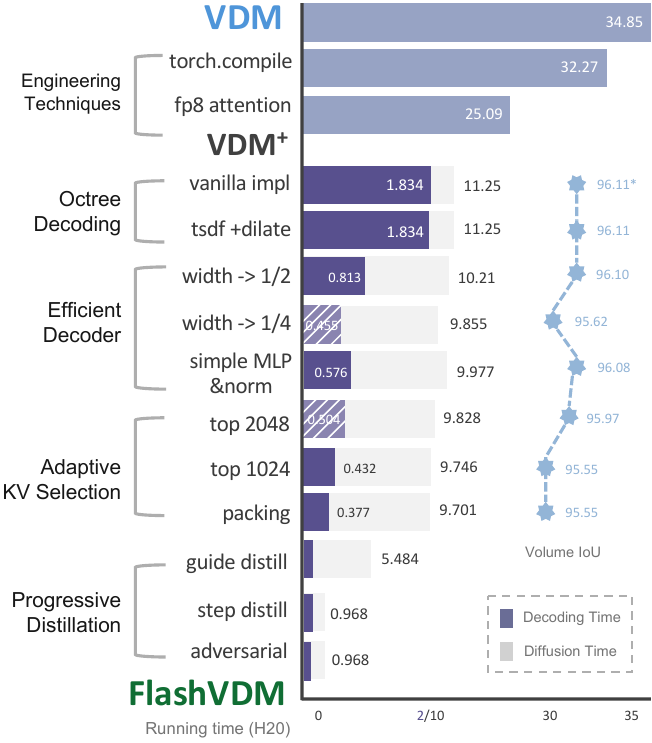}
   \caption{Step-by-step case study of \shortname acceleration techniques and their effects on inference time and IoU. }
   \foobar{-4mm}
   \label{fig:step-by-step-time}
\end{figure}

Specifically, instead of directly decoding the SDF volume at the target resolution (\eg, 384), we begin by decoding at a smaller resolution (\eg, 75), which provides a coarse SDF volume. Using this coarse volume, we can identify which voxels intersect the shape surface based on the following criterion:
\begin{equation}
f(x,y,z) = 
\begin{cases} 
1 & \text{if } \exists (i,j,k) \in N(x,y,z) \\
& \text{such that } f(i,j,k) \neq f(x,y,z), \\
0 & \text{otherwise},
\end{cases}
\label{eq:intersect}
\end{equation}
where $N(x,y,z)$ represents the set of adjacent voxels to $(x,y,z)$ and $f(x,y,z)=1$ indicates the voxel intersects with the surface. 
In short, a voxel does not intersect with the surface only if all its adjacent voxels are either inside or outside. For voxels that do intersect the surface, we subdivide them into higher resolutions and compute the SDF values through cross-attention once more. This process is repeated until we reach the target resolution. Since the surface shape is highly sparse within the volume, the number of effective queries is significantly reduced.

Naively, previous steps could be considered a type of octree decoding~\cite{saito2019pifu}. However, we find that it produces artifacts and holes when applied to VDM in practice. To prevent degradation, we need to consider some corner cases. First, when the resolution is low, for very thin meshes, two adjacent voxels may lie on opposite sides of the mesh surface, both having the same sign, as shown in \cref{fig:ovd_corner_cases} (a). This causes the missing voxels for resolution increment, leading to holes in the surface.
To address this, we use tSDF as supervision during VAE training to help determine whether a voxel is near the mesh surface. Then, during \svdname, we append all voxels under a certain tSDF threshold~(TR). With this strategy, the lowest resolution in \svdname can be reduced to $w+u*2$ where $w$ is the minimal mesh width and $u$ is the tSDF TR. Second, failure cases exist even with tSDF augmentation as shown in \cref{fig:ovd_corner_cases} (b,c), \eg, uneven distribution of surface between two adjacent voxels and \etc. Therefore, we perform a dilation operation after identifying the intersected voxels to prevent unexpected missing points. We leave the choice of tSDF value and more details to \cref{appendix:ovd}.

\begin{figure}[t]
  \centering
  \includegraphics[width=\linewidth]{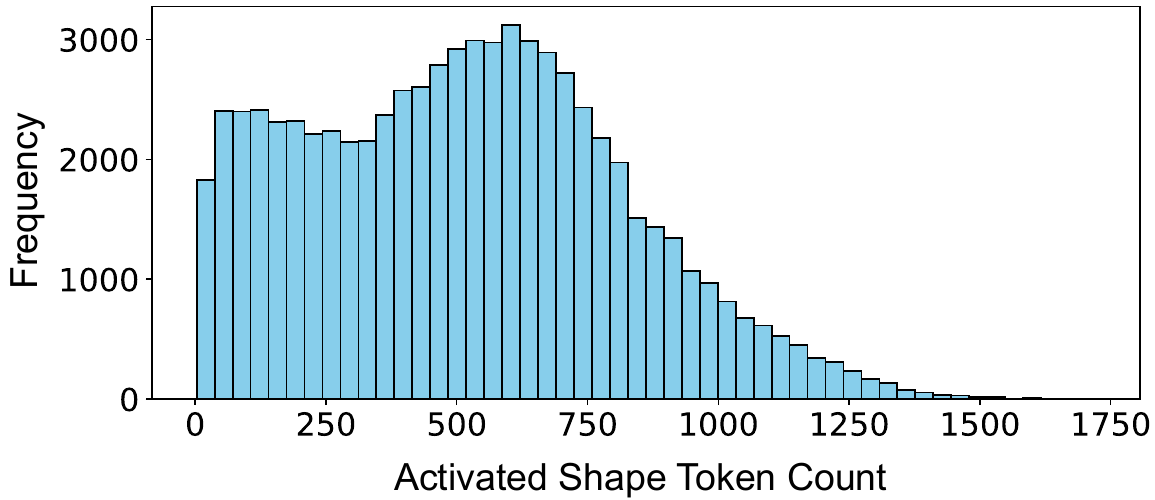}
   \caption{Histogram of activated token counts with non-zero attention of 70,800 regions from 250 cases.}
   \foobar{-4mm}
    \label{fig:histogram_token_counts}
\end{figure}

\begin{figure}[t]
  \centering
  \includegraphics[width=\linewidth]{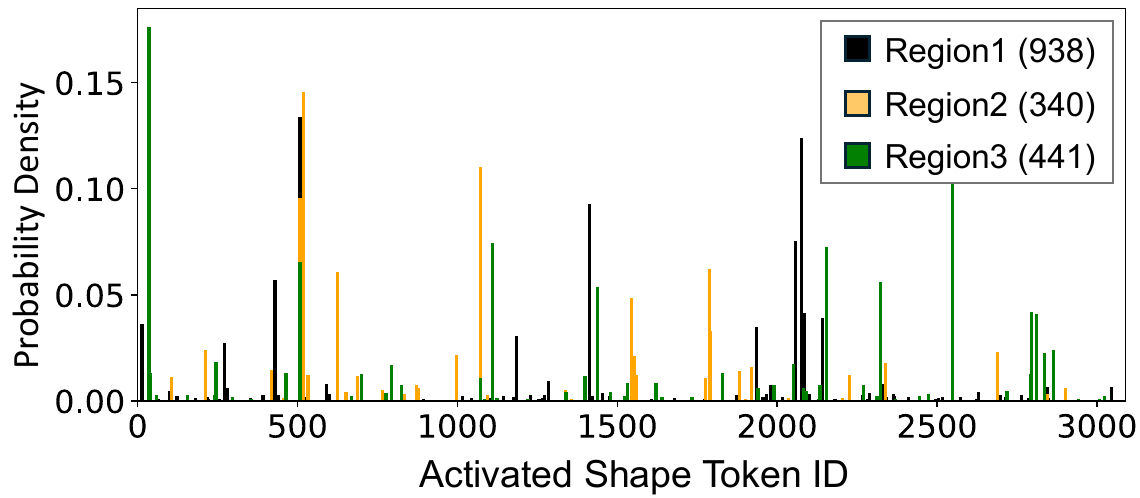}
   \caption{Normalized histogram of activated shape tokens with non-zero attention at different regions. Zoom in for a better view. The numbers in the legend indicate the number of activated tokens.}
   \foobar{-4mm}
   \label{fig:locality}
\end{figure}

\textbf{Adaptive KV Selection.} 
For each position in volume decoding, a cross-attention operation is performed between the queries and key-value pairs (computed from shape latents). Previously, we significantly reduced the number of queries. To further minimize computational overhead, we introduce adaptive KV selection, which reduces the number of key-value pairs. Our method leverages the observation that the correlation between spatial queries and shape latents exhibits strong locality -- adjacent spatial points tend to attend to similar latent tokens, with the attention concentrated on a small subset of tokens (denoted as activated tokens). To illustrate, we show two histograms in \cref{fig:histogram_token_counts,fig:locality}. As seen, most regions in 3D volumes attend to at most 1,000 tokens -- one-third of the original 3,072, and different regions focus on different set of activated tokens. Moreover, our statistic reveals that the average activated token count for a single query is around only 10, which also indicates strong locality in shape latents. 

The proposed adaptive KV selection is training-free and able to be combined with \lvdname. For simplicity, we first show the algorithm with vanilla volume decoding.
In this case, we first divide the entire volume into smaller sub-volumes. Then, within each sub-volume, we uniformly sample a small set of queries to compute their attention scores relative to the keys. To select most correlated key-value pairs, we could simply average the attention score of all queries and select the TopK, or we could merge the TopN of each query to obtain TopM, where N is a small number (\eg, 50) that can include most activated tokens with non-zero attention. The latter one is slightly better but also slightly slower~(5\%). For other queries within the same sub-volume, we only use the Top-K/M key-value pairs for attention computation. Since the attention scores are estimated using a small set of queries and the number of activated shape latents are much smaller, the overall computation can be reduced by 34\%, as shown in \cref{fig:step-by-step-time}.

The combination with \lvdname needs some extra consideration to ensure performance.
The core obstacle lies in the requirement that subvolume should be small to maintain locality. Within \svdname, the effective queries could be even fewer in a subvolume.
Thereafter, naive implementation that process each subvolume one-by-one would not maximize GPU utilization. As a remedy, we design a packing operation, which precompute the target queries in each volume, pack queries of different subvolume, and process them together. More details on the analysis and the implementation lives in  \cref{appendix:aks_impl}.

\textbf{Efficient Decoder Design.}
Once we have reduced the QKV in CA, further acceleration of the attention operation itself becomes challenging algorithmically. Thereby, our next approach focuses on optimizing the decoder network design. Most existing vectorset decoders~\cite{zhang20233dshape2vecset,zhao2024michelangelo,li2024craftsman} share nearly identical architectures, typically consisting of several self-attention (SA) layers and a cross-attention layer, as shown in \cref{fig:vdm}, with SA layers evaluated only once. Therefore, our goal is to simplify the final CA layer as much as possible, which we believe could be replaced with a simpler one as it only needs to classify the inside/outside status of each position. To this end, we reduce the network width, decrease the MLP expansion ratio, and remove several LayerNorm layers. Through careful ablation studies, we found that these reductions have minimal impact on reconstruction quality while boosting speed by 3.2\texttimes.

\subsection{Progressive Flow Distillation}
\label{sec:pfd}

\begin{figure}[t]
  \centering
  \includegraphics[width=\linewidth]{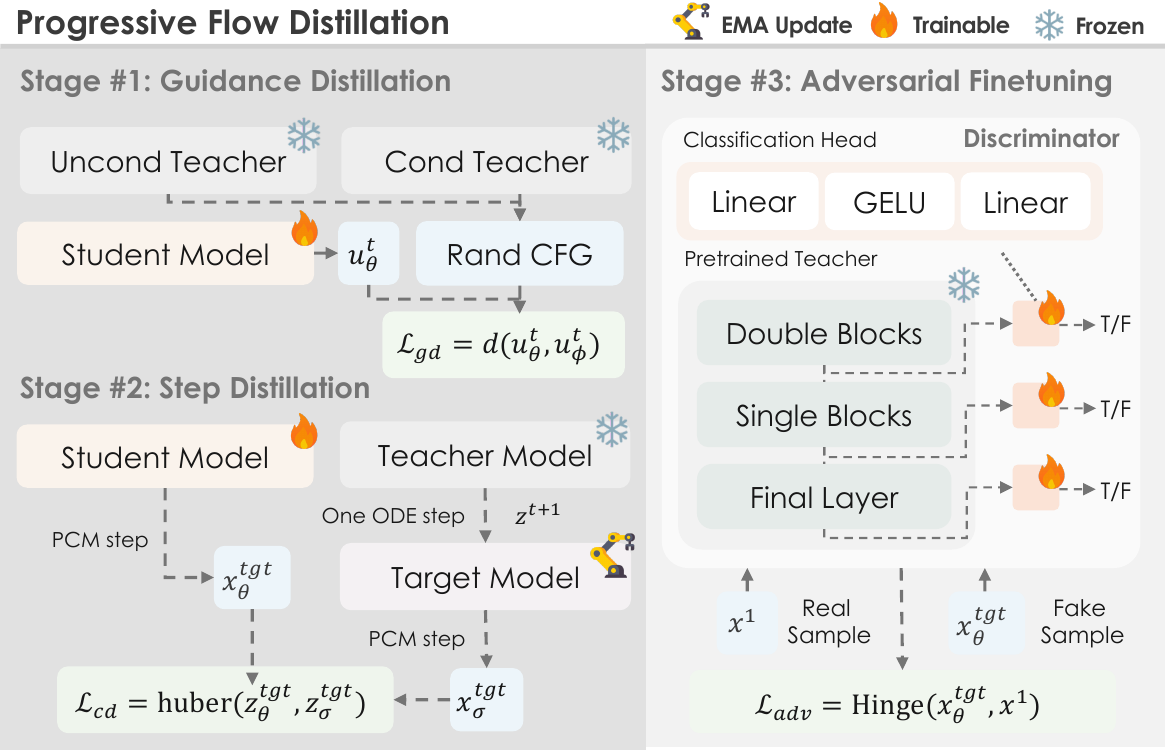}
   \caption{Training pipeline for Progressive Flow Distillation.}
   \label{fig:pfd}
\end{figure}

Similar to image and video generation~\cite{rombach2022high, li2024hunyuandit, kong2024hunyuanvideo}, we can reduce the sampling steps of VDM through distillation~\cite{luo2023latent, zhu2024accelerating}. 
Initially, we experimented with PCM~\cite{wang2025phased}, but we found it hard to yield satisfactory results on VDM.
To address this, we propose progressive flow distillation as shown in \cref{fig:pfd}, systematically identifying and refining key design choices for effective distillation.

\textbf{Consistency Flow Distillation (CFD).} Consistency Distillation (CD)~\cite{song2023consistency} is a well-known distillation method for reducing the diffusion sampling steps, though its application to 3D diffusion has yet to be explored.
The core of CD is to enforce a consistency property -- any point along the ODE trajectory maps to the same target point. To achieve this, CD uses a time difference method, where a starting point $\mathbf{x}_{t_n}$ is first selected for the target prediction of the student model $\mathbf{x}_0^{t_n}=\boldsymbol{f}_{\boldsymbol{\theta}}\left(\mathbf{x}_{t_{n}}, t_{n}\right)$. For flow-based models, \eg, Hunyuan3D-2~\cite{zhao2025hunyuan3d}, the Euler solver could be employed with a large discretization step $t_n-t_0$ to compute the target $\mathbf{x}_0^{t_n}$. 
For GT target, the teacher model predicts the next point $\hat{\mathbf{x}}_{t_{n+1}}^{\boldsymbol{\phi}}$ by running one discretization step of an ODE solver, after which the student model predicts the GT target at next timestep $x_0^{t_{n+1}}=\boldsymbol{f}_{\boldsymbol{\theta}}\left(\hat{\mathbf{x}}_{t_{n+1}}^{\boldsymbol{\phi}}, t_{n+1}\right)$. The consistency property is enforced by the following loss,
\begin{equation}
\mathcal{L}_{cfd}(\boldsymbol{\theta}) := \\
\mathbb{E}\left[d\left(\boldsymbol{f}_{\boldsymbol{\theta}}\left(\mathbf{x}_{t_{n}}, t_{n}\right), \boldsymbol{f}_{\boldsymbol{\theta^{-}}}\left(\hat{\mathbf{x}}_{t_{n+1}}^{\boldsymbol{\phi}}, t_{n+1}\right)\right)\right],
\end{equation}
with a boundary constraint $f(\mathbf{x}_{\epsilon}, \epsilon)=\mathbf{x}_{\epsilon}$, where $d$ is a distance function and $\theta^{-}$ is a regular copy of student model named as target model. 

\textbf{Stablizing Target Model.} 
In practice, directly applying the distillation method described above does not yield satisfactory results, as shown in \cref{fig:distill_ablation} (w/o GD distill). Our analysis reveals that the main challenge lies in the stability of the target model. During training, the student model continually tries to mimic the output of the target model, making it highly sensitive to changes in the target. As a result, we observe significant fluctuations in the predictions throughout the training process. To stabilize the target model, we introduce a progressive training strategy, in which 1) we first perform guidance distillation and initialize the student model as a distilled version. This approach differs from distillation in image models, where guidance distillation can be applied simultaneously with step distillation without issues. We hypothesize that this gap arises from the differences between 2D and 3D models and the more complex optimization landscape in 3D models. 2) Besides, we find that the EMA update of the target model is crucial for stabilizing VDM, which contrasts with a recent study on 2D generation as well \cite{song2023improved}. 3) Moreover, we replace commonly used L2 loss with Huber loss as it is less sensitive to outliers, which makes training more stable. 4) Finally, we introduce a multi-stage-multi-phase consistency distillation strategy based on PCM~\cite{wang2025phased}, in which a single-phase finetuning is performed after five phases pretraining.

\begin{figure}[t]
  \centering
  \includegraphics[width=\linewidth]{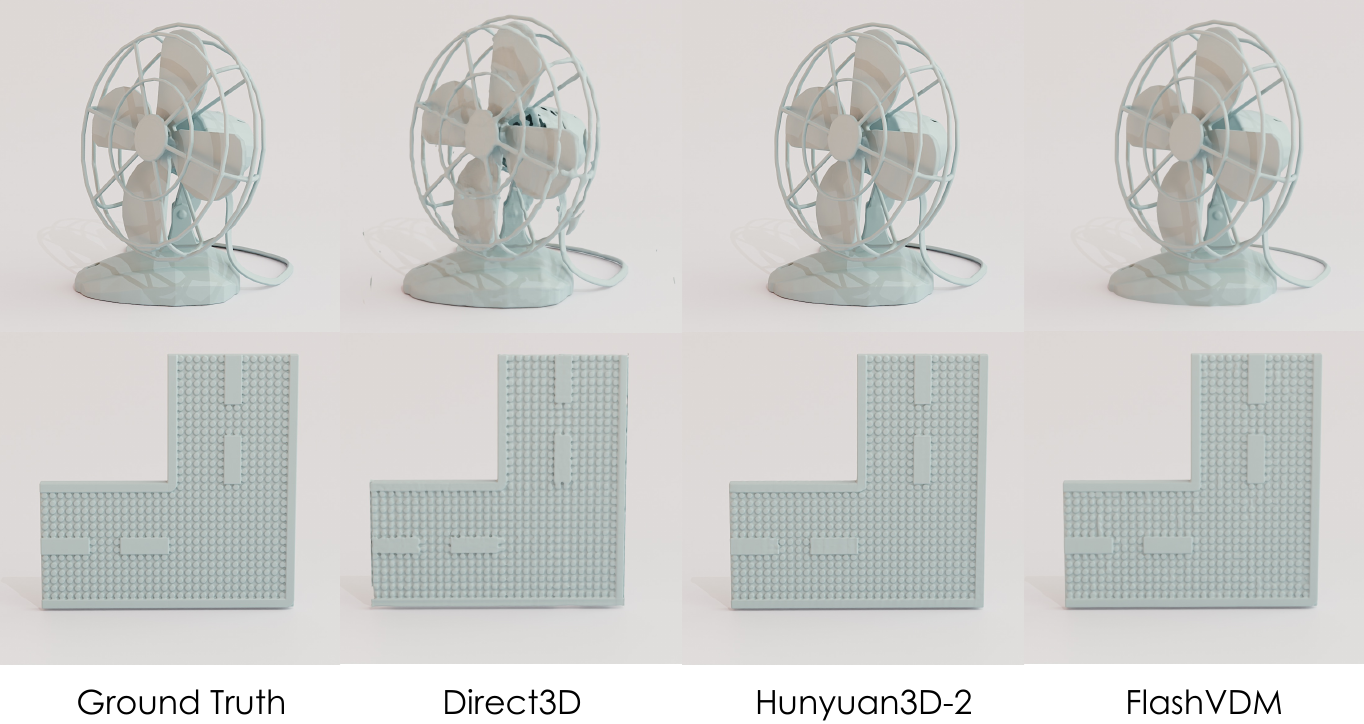}
   \caption{Visual comparison of shape reconstruction methods.}
   \label{fig:vae_reuslt}
\end{figure}
\input{tables/cmp_vae}

\begin{figure*}[t]
  \centering
  \includegraphics[width=\linewidth]{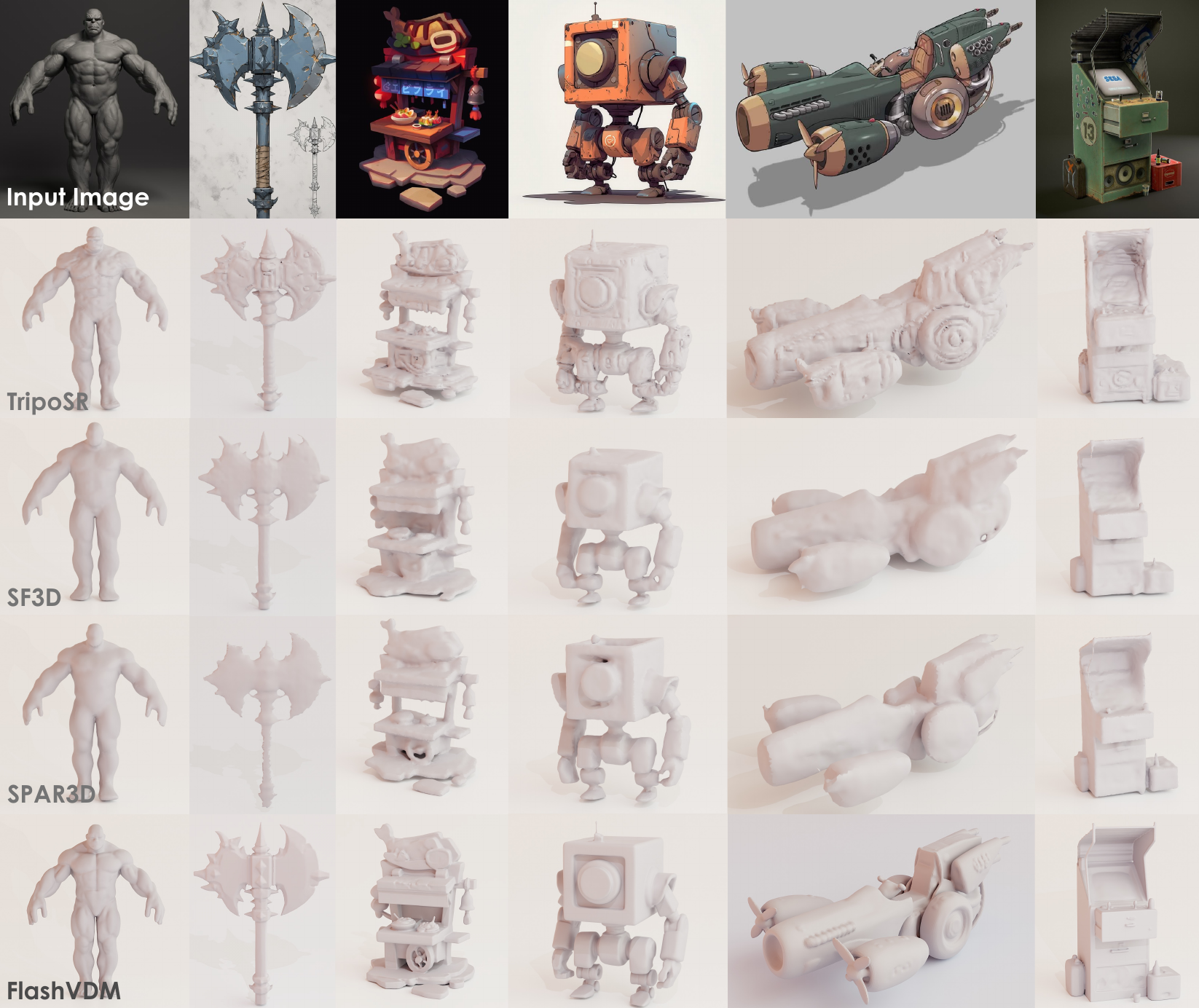}
   \caption{Visual comparison of image-to-3D generation between the proposed FlashVDM and other fast 3D generation methods.}
   \foobar{-4mm}
   \label{fig:dit_result_fast}
\end{figure*}

\textbf{Aligning Real Data with GAN.} 
With the proposed improvements, our CFD is able to reduce the sampling steps to just 5 while still producing decent results. However, we found that in some cases, the mesh quality is difficult to match the teacher’s outputs, which might be attributed to the self-distillation nature of the consistency model. To address this, we seek to further enhance performance by incorporating supervision from ground-truth 3D data through adversarial training~\cite{Goodfellow2014GAN}.
Specifically, we initialize the generator using the model distilled in the previous stage. 
Inspired by previous works~\cite{sauer2024fast, xu2023ufogen}, our discriminator operates directly in latent space, eliminating the need for an expensive decoding process, and leverages the intermediate features of the pretrained diffusion model. 
As shown in \cref{fig:pfd}, we first extract token sequences from specified attention layers, and then apply independent discriminator heads. Unlike prior works~\cite{sauer2024fast, wang2025phased, xu2023ufogen, yin2024improved} that use noised latents, we did not observe significant improvements with noise, so we opt to use ground-truth latents for simplicity.
The discriminator is trained with a hinge adversarial loss~\cite{lim2017geometric} as,
\begin{equation}
\mathcal{L}_{\mathrm{adv}}^{}(\theta, \gamma) = \text{ReLU}(1+\mathcal{D}_\gamma(\mathbf{x}_0)) + \text{ReLU}(1-\mathcal{D}_\gamma(\mathbf{x}_0^{t_{n}})),
\end{equation}
which distinguishes between ground truth latent $\mathbf{x}_0$ and latent produced by our generator $\mathbf{x}_0^{t_{n}} = f_\theta(z_{t_n}, t_n)$.
Following \cite{yin2024improved,wang2025phased}, the final objective is a combination of the adversarial objective and consistency loss $\mathcal{L} = \mathcal{L}_{cfd} + \lambda\mathcal{L}_{adv}$ where $\lambda$ is set to 0.1.

%% file: tables/cmp_vae.tex
\begin{table}
\centering
\small
\begin{tabular}{rccc}
\hline
     & \textbf{V-IoU($\uparrow$)} & \textbf{S-IoU($\uparrow$)} &  \textbf{Time(s$\downarrow$)}   \\ \hline
3DShape2VecSet~\cite{zhang20233dshape2vecset}   & 87.88\%    & 84.93\%     & 16.43 \\ 
Michelangelo~\cite{zhao2024michelangelo}      & 84.93\%    & 76.27\%     & 16.43     \\    
Direct3D~\cite{wu2024direct3d}       & 88.43\%    & 81.55\%     &  3.201    \\ \hline   
Hunyuan3D-2~\cite{zhao2025hunyuan3d}-1024  & 93.60\%    &  89.16\%    & 16.43    \\ 
\rowcolor{graycolor}
$\llcorner$ with \shortname  &   91.90\%  & 88.02\%    & 0.382  
\\    
Hunyuan3D-2~\cite{zhao2025hunyuan3d}-3072  &    96.11\% &     93.27\% & 22.33    \\ 
\rowcolor{graycolor}
$\llcorner$ with \shortname  &   95.55\%  &    93.10\% & 0.491 
\\    
\hline
\end{tabular}

\caption{Numerical comparisons of shape reconstruction methods. }
\label{tab:recon}
\end{table}

%% file: sec/5_experiment.tex
\section{Experiments}

\begin{figure}[t]
  \centering
  \includegraphics[width=\linewidth]{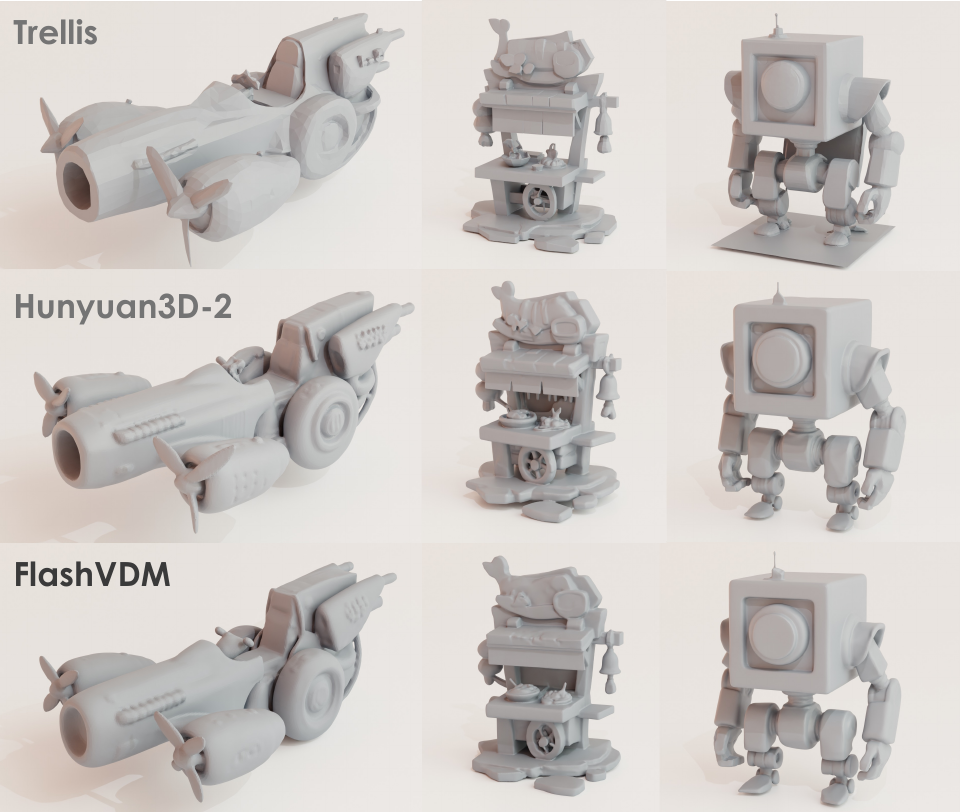}
  \foobar{-5mm}
   \caption{Visual comparison of image-to-3D generation between the proposed FlashVDM and other 3D diffusion methods.}
   \foobar{-2mm}
   \label{fig:dit_result}
\end{figure}

In this section, we apply the proposed \shortname to \hyshape~\cite{zhao2025hunyuan3d}, which is currently a state-of-the-art open-source VDM. We evaluate our approach in terms of both VAE reconstruction and diffusion generation. We also provide ablation studies of the proposed techniques.

\subsection{Reconstruction}

\textbf{Metrics.} We utilize the volume and surface IoU metric to assess the impact of our acceleration techniques on VAE reconstruction performance. 
The running time is measured at the resolution of 380.

\textbf{Comparison.} We compare our method with three competing method, \ie, 3DShape2VecSet~\cite{zhang20233dshape2vecset}, Michelangelo~\cite{zhao2024michelangelo}, and Direct3D~\cite{wu2024direct3d}. We evaluate two resolutions, \ie, 1,024 and 3,072, for our fast version and the base model Hunyuan3D-2~\cite{zhao2025hunyuan3d}.  The numerical comparison is shown in \cref{tab:recon}. It demonstrates that our method outperforms all competing methods and preserve the quality of base model with less than 1\% IoU drop while obtaining over 45\texttimes ~speedup. \cref{fig:vae_reuslt} shows the visual comparison, which also indicates minimal degradation in quality. 

\begin{figure}[t]
  \centering
  \includegraphics[width=\linewidth]{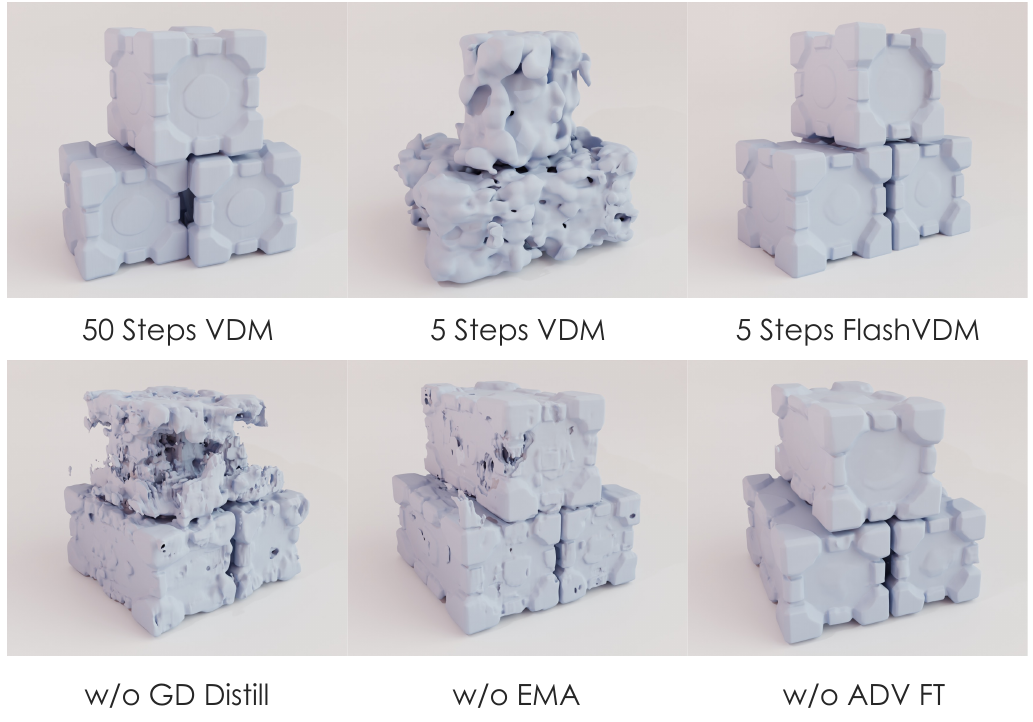}
  \foobar{-6mm}
   \caption{Ablation study of our progressive flow distillation.}
   \foobar{-2mm}
   \label{fig:distill_ablation}
\end{figure}

\begin{figure}[t]
  \centering
  \includegraphics[width=\linewidth]{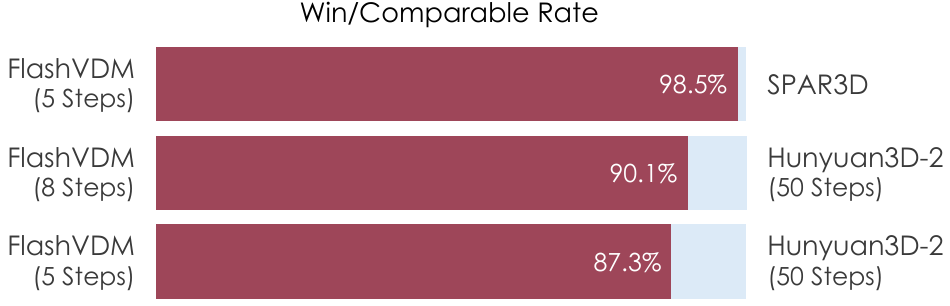}
  \foobar{-4mm}
   \caption{User study of \shortname against different methods.}
   \label{fig:user_study}
\end{figure}

\subsection{Generation}

\textbf{Metrics.} To evaluate shape generation performance, we adopt ULIP-I~\cite{xue2023ulip} and Uni3D-I~\cite{zhou2023uni3d} to compute the similarity between the generated mesh and input images.

\textbf{Comparison.} Three latest fast 3D generation methods, \ie, TripoSR~\cite{tochilkin2024triposr}, SF3D~\cite{boss2024sf3d}, and SPAR3D~\cite{huang2025spar3d}, and two state-of-the-art methods \hyshape~\cite{zhao2025hunyuan3d} and Trellis~\cite{xiang2024structured} are selected for comparison. The numerical comparison is shown in \cref{tab:dit_result} and the visual comparison is shown in \cref{fig:dit_result_fast,fig:dit_result}. It can be observed that our method retain most of capability of the base model, while achieves the best results with a large margin against other fast methods.

\input{tables/cmp_dit}
\input{tables/ablation_vae}

\textbf{User Study.} We include results of three user studies in \cref{fig:user_study}. For comparison between FlashVDM and SPAR3D~\cite{huang2025spar3d}, we ask participants to determine which one is better. For comparison between FlashVDM and Hunyuan3D-2~\cite{zhao2025hunyuan3d}, we ask whether there is a huge difference.  As seen, FlashVDM is preferred over SPAR3D~\cite{huang2025spar3d} in almost all testcases, and FlashVDM is comparable to its base model Hunyuan3D-2~\cite{zhao2025hunyuan3d} at 5 steps and the preference rate increase with more steps.

\subsection{Ablation Study}
\label{sec:ablation}

\textbf{Effectiveness of Lightning Vecset Decoder.} The results of step-by-step ablation is shown in \cref{tab:recon_ablation}, in which we add each component incrementally. It can be observed that \svdname provides a 10x speedup with no degradation in quality. The efficient decoder achieves an extra 3x speedup with almost no degradation, while adaptive KV selection results in a 30\% speedup with minimal degradation.

\textbf{Effectiveness of Progressive Flow Distillation.} 
We present an ablation study to demonstrate the effectiveness of progressive training strategies. The visual comparison is shown in \cref{fig:distill_ablation}. As seen, the original VDM completely fails at 5 steps, while FlashVDM achieves results comparable to the original 50-step VDM. Additionally, we observe that distillation fails without guidance distillation as a warm-up step, and performance degrades without the use of EMA. Finally, adversarial fine-tuning is shown to help generate smoother and more accurate shapes. More detailed ablation can be found in the \cref{sec:appendix_distill}.

%% file: tables/cmp_dit.tex
\begin{table}
\centering
\small
\begin{tabular}{rccc}
\hline
 & \textbf{ULIP-I($\uparrow$)} & \textbf{Uni3D-I($\uparrow$)} & \textbf{Time(s$\downarrow$)} \\ \hline
TripoSR~\cite{tochilkin2024triposr} & 0.0642  & 0.1425 & 0.958 \\
SF3D~\cite{boss2024sf3d}    & 0.1156  & 0.2676  & 0.212 \\
SPAR3D~\cite{huang2025spar3d}       & 0.1149 & 0.2679  & 1.296 \\
\hline
Trellis~\cite{xiang2024structured}                   & 0.1267 & 0.3116 & 7.334 \\
Hunyuan3D-2~\cite{zhao2025hunyuan3d}                   & 0.1303 & 0.3151 & 34.85 \\ 
\rowcolor{graycolor}
$\llcorner$ with \shortname  &  0.1260    &  0.3095  & 1.041     \\ 
\hline
\end{tabular}
\caption{Numerical comparisons of shape generation methods.}
\label{tab:dit_result}
\end{table}

%% file: tables/ablation_vae.tex
\begin{table}
\centering
\small
\begin{tabular}{rccc}
\hline
     & \textbf{V-IoU($\uparrow$)} & \textbf{S-IoU($\uparrow$)} &  \textbf{Time(s$\downarrow$)}   \\ \hline
VAE Baseline  &    96.11\%  &   93.27\% &  22.33   \\
+ \scvdname  &    96.11\%  &    93.27\% & 2.322   \\ 
+ Efficient Decoder  &   96.08\%  &    93.13\% & 0.731   \\ 
+ Adaptive KV Selection &     95.55\%  &    93.10\% & 0.491   \\
\hline
\end{tabular}
\caption{Step-by-step ablation of our lightning vecset decoder. }
\foobar{-4mm}
\label{tab:recon_ablation}
\end{table}

%% file: sec/6_conclusion.tex
\section{Conclusion}
In this work, we introduce \textit{\shortname}, a general framework for accelerating a pretrained VDM~\cite{zhao2025hunyuan3d, li2025triposg}. Our framework encompasses not only a progressive flow distillation method for distilling VDM into a few-step generator, but also several training-free inference techniques and an efficient, lightweight vecset decoder that significantly reduces the FLOPs of decoding. We apply our framework to the state-of-the-art image-to-3D VDM, \hyshape~\cite{zhao2025hunyuan3d}, obtaining \hyshape Turbo. Our evaluation demonstrates that \shortname excels in both reconstruction and generation, achieving \vaespeedup and \ditspeedup speedups, respectively. To the best of our knowledge, \shortname is the first work to push large-scale shape generation into the millisecond range, which opens up new possibilities for interactive applications of 3D generative models.

%% file: sec/x_appendix.tex
\section{Implementation Details}

\textbf{Decoder Finetuning.}
The efficient vecset decoder illustrated in \cref{sec:lvd} is fine-tuned by freezing the vecset encoder. In \cite{zhao2025hunyuan3d}, the decoder is made up of 8 self-attention layers and 1 cross-attention layer. Since our design only alters the cross-attention layer, we initialize self-attention layers as before. Both self- and cross-attention layers are trained during the finetuning. We use a constant learning rate of $1\times 10^{-4}$ and a batch size of 256. The decoder could quickly converge to a pretty good one with 300k steps, but we find longer training to 800k steps converges better, leading to nearly identical performance to the original one. 

\textbf{Diffusion Distillation.}
The batch size is always 256 for different stages in progressive flow distillation. Following~\cite{luo2023latent, meng2023distillation}, the guidance distilled model is conditioned on the guidance strength $w$, which is injected into the diffusion backbone with a similar approach as timestep. During training, $w$ is randomly select from $w\sim U[2,8]$. The learning rate is set to $1\times 10^{-6}$. The model is trained with 20k steps. For step distillation, we set $\lambda$ of huber loss to $1\times 10^{-3}$, and the guidance strength is set to a constant of 5.0. Following~\cite{luo2023latent}, we also use the skipping-step technique with $k=10$. We utilize multiphase~\cite{wang2025phased} techniques to train the model for 20k steps with 5 phases and a learning rate of $1\times 10^{-6}$ and then finetune the model for 8k steps with a learning rate of $1\times 10^{-7}$. The EMA decay rate is set to $0.999$. For adversarial fine-tuning, we keep the distillation loss of the previous stage and set the adversarial loss weight to 0.1. The learning rate is set to $1\times 10^{-7}$ for generator and $1\times 10^{-6}$ for discriminator. We train 5k steps for this stage.

\section{Details of \vdname}
\label{appendix:ovd}

\begin{figure}[t]
  \centering
  \includegraphics[width=\linewidth]{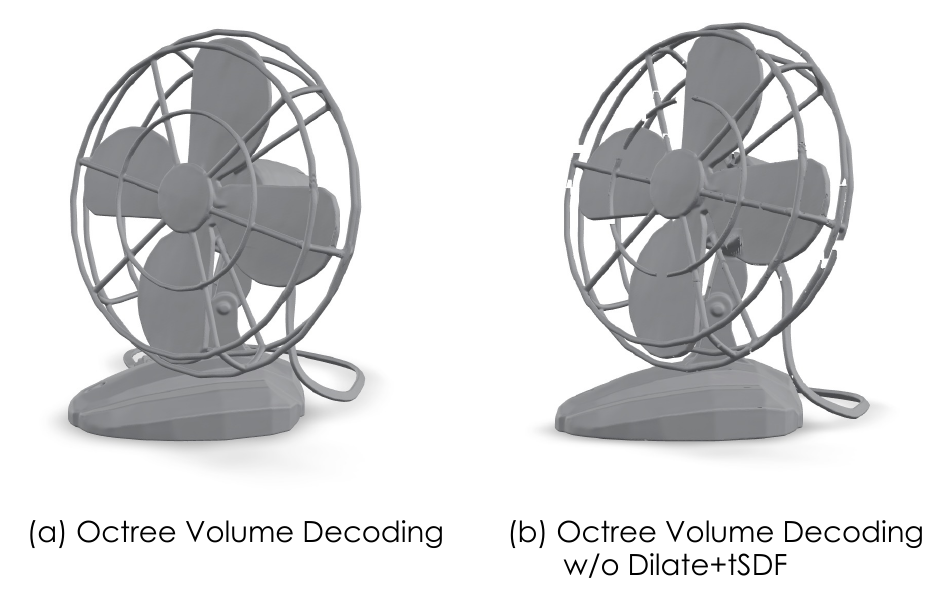}
   \caption{Comparison of reconstruction results with and without dilate and tSDF strategy for \lvdname.}
   \label{fig:ablation_dilate}
\end{figure}

\textbf{Effect of Dilate and tSDF.} \cref{fig:ablation_dilate} compares the reconstruction with and without dilate and tSDF strategy. It can be seen that dilate+tSDF is mandatory for reconstructing complete mesh without holes.

\textbf{Implementation and Practical Consideration.}
The overall pseudocode for \lvdname is shown in \cref{algo:decode}. In practice, we set the tSDF threshold $\eta=0.95$ and the isosurface threshold $\gamma=0.0$. The dilate operation is implemented using a 3D convolution with a kernel size of 3. At the final resolution, the total number of points increases significantly, thus the \texttt{FindNear} operation would introduce many redundant points. To address this, we omit the \texttt{FindNear} operation while \texttt{Expand} twice, striking a balance between speed and quality. Practically, we find this strategy has minimal impact on the overall quality while speeding up slightly.

\begin{algorithm}[t]
    \caption{\vdname.}
    \begin{algorithmic}[1]
        \Require An implicit function $f(\mathbf{p})$ that evaluates the SDF at position $\mathbf{p}$. Target resolution $\mathbf{r}$. Shape latents $Z$, tSDF threshold $\eta$, isosurface threshold $\gamma$.
        \Ensure A signed distance field (SDF) $S\in \mathcal{R}^{r\times r\times r}$. 
        \State $R = \text{GetResolutions}(r) $ \Comment{List of cascade resolution.}
        \State $P_{R[0]} = \text{GenGridPoints}(R[0])$
        \State $S_{R[0]} = \text{QueryField}(P_{R[0]},Z)$
        \For{$i = 1$ \textbf{to} $\text{len}(R)$}
            \State $\hat{P}_{R[i]} = \text{FindIntersect}(S_{R[i-1]}, \gamma)$
            \State $\hat{P}_{R[i]} \mathrel{{+}{=}} \text{FindNear}(S_{R[i-1]}, \eta)$
            \State $\hat{P}_{R[i]} = \text{Dilate}(\hat{P}_{R[i]})$
            \State $P_{R[i]} = \text{Expand}(\hat{P}_{R[i]})$
            \State $S_{R[i]} = \text{QueryField}(P_{R[i]},Z)$
        \EndFor \\
        \Return $S_{R[-1]}$
    \end{algorithmic}
    \label{algo:decode}
\end{algorithm}

\section{Details of Adative KV Selection}
\label{appendix:aks}

\subsection{Analysis of Locality.}

\begin{figure}[t]
  \centering
  \includegraphics[width=\linewidth]{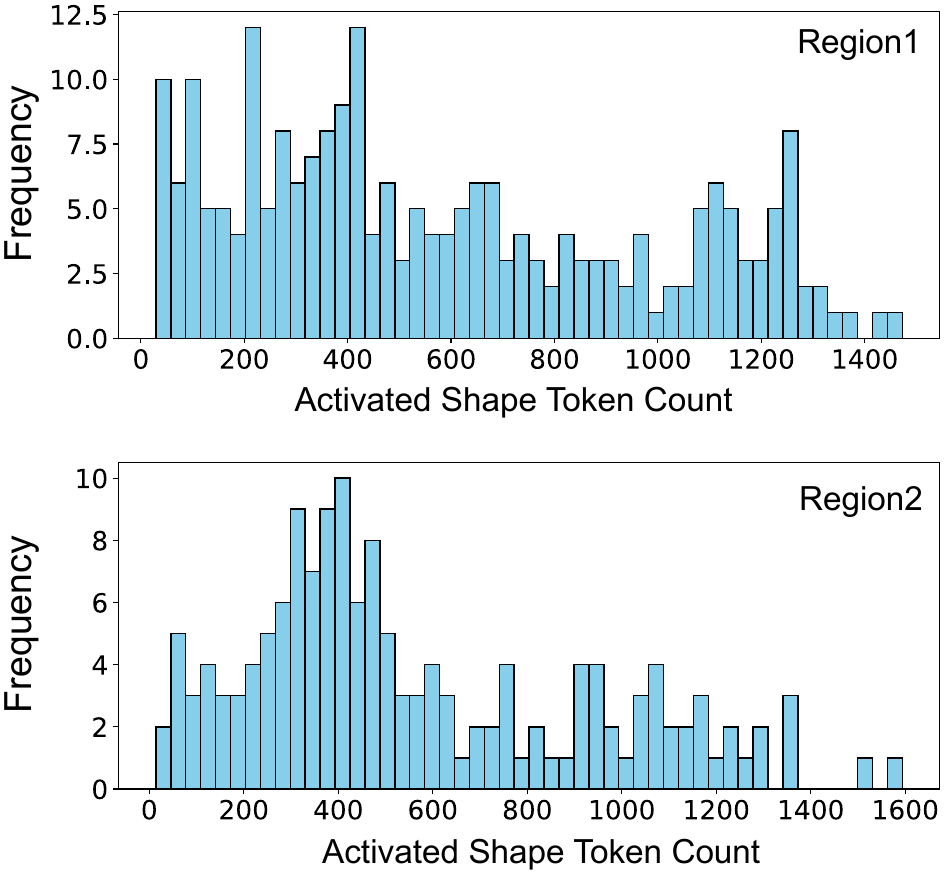}
   \caption{Histogram of activated token counts within different regions, measured with 300 cases.}
    \label{fig:histogram_token_counts_per_region}
\end{figure}

\textbf{Activated Tokens Across Different Cases in the Same Region.} \cref{fig:histogram_token_counts_per_region} presents a histogram of the activated shape token count across 300 different test cases. As observed, different regions and cases activate different sets of tokens. This suggests that locality is case-dependent rather than region-dependent. In other words, the same region in different cases does not consistently share a similar set of activated tokens.

\begin{figure}[t]
  \centering
  \includegraphics[width=\linewidth]{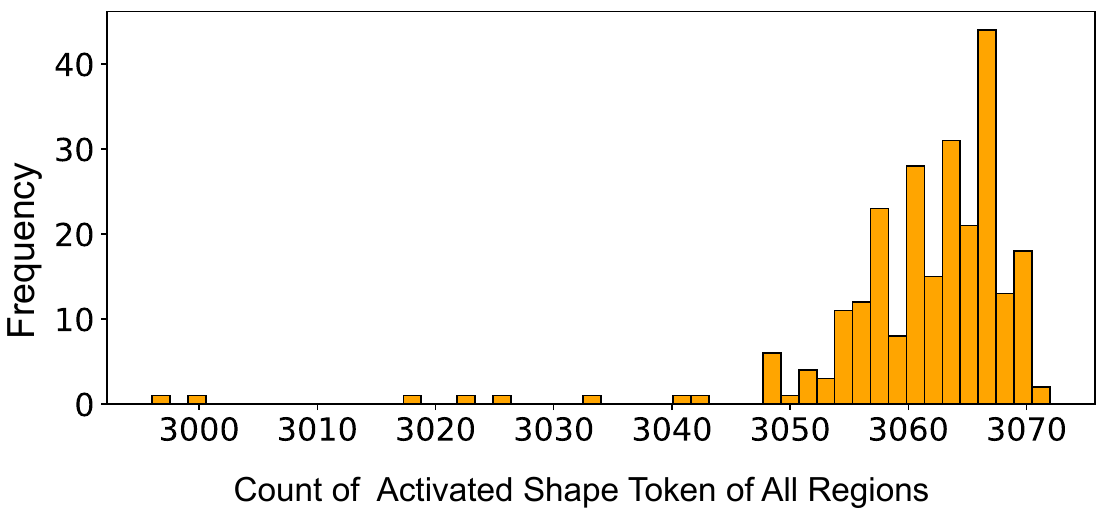}
   \caption{Histogram of the number of total activated token within all regions, measured with 200 cases.}
    \label{fig:histogram_total_count}
\end{figure}

\textbf{Distribution of Activated Tokens Within a Case.} \cref{fig:histogram_total_count} shows the histogram of the total number of activated tokens within a case, based on 200 cases. It can be observed that most cases contain over 3000 tokens, with a maximum of 3072 tokens. This further confirms that the phenomenon of having fewer activated tokens per region is due to token locality, rather than token redundancy.

\begin{figure}[t]
  \centering
  \includegraphics[width=\linewidth]{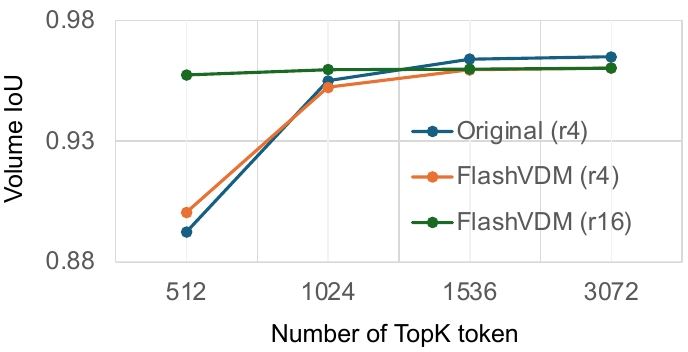}
   \caption{The graph shows the relationship between volume IoU and the number of TopK tokens. r4 denotes the volume is divided into $4^3$ subvolumes, and r16 denotes  $16^3$ subvolumes. }
   \label{fig:topk_iou}
\end{figure}

\textbf{IoU Changes with Respect to TopK Tokens.} \cref{fig:topk_iou} shows the relationship between volume IoU and the number of TopK tokens, with all methods utilizing \lvdname. The results for Original and FlashVDM differ due to the use of the efficient decoder. It can be observed that FlashVDM(r4) closely matches the curve of Original(r4), suggesting that our efficient decoder design preserves most of the reconstruction ability. Additionally, we notice that r16 performs significantly better than r4, highlighting the strong locality of attention between queries and shape tokens. Higher resolution corresponds to smaller subvolumes, resulting in improved locality. Interestingly, r16 maintains a similar IoU even with just 16\% (512/3072) of the tokens.

\subsection{Implementation} 
\label{appendix:aks_impl}

\textbf{Combination with \vdname.} 
In \cref{sec:lvd}, we briefly introduce the combination of Adaptive KV Selection (AKVS) and \lvdname. Here, we provide a more detailed explanation of the implementation and background. AKVS can be naively implemented as shown in \cref{algo:kvs}. The algorithm consists of four main steps: sampling queries, computing the mean attention score, selecting Top-K, and performing attention. Since the attention score is computed from the sampled queries, we need to feed the queries subvolume by subvolume, with queries being spatially close to one another, to keep locality.

For the original volume decoding, this can be easily achieved by changing the chunk-splitting method to a subvolume-splitting method, as the original method also uses chunk-splitting to reduce memory requirements. However, to maintain locality, the chunk size must be much smaller, which may be too small for efficient GPU acceleration. Additionally, with \lvdname, the number of queries in each subvolume can be even smaller. To address this, we propose to pre-divide the subvolume and concatenate all queries of each subvolume. Instead of processing each subvolume sequentially, we concatenate multiple subvolumes and process them in parallel until cross-attention is reached. This approach helps reduce the running time for MLP and other linear layers in the decoder.

\begin{algorithm}[t]
  \caption{Adaptive KV Selection.}
  \begin{algorithmic}[1]
      \Require Query $Q\in \mathcal{R}^{N\times D}$, Key $K\in \mathcal{R}^{M\times D}$, and Value $V\in \mathcal{R}^{M\times D}$ of cross attention, the number of queries $n\ll N$ for estimating TopK correlated KV.
      \Ensure Attention result $O\in \mathcal{R}^{N\times D}$.
      \State $\hat{Q} = \text{Sample}(Q)$, $\hat{Q}\in \mathcal{R}^{n\times D}$
      \State M = $\text{Mean}(\hat{Q}\times \hat{K}^T)$, $M\in \mathcal{R}^{n\times M}$
      \State $\hat{K}, \hat{V} = \text{TopK}(M, K,V)$, $\hat{K}, \hat{V}\in \mathcal{R}^{k\times D}$
      \State $O=\text{Attention}(Q, \hat{K}, \hat{V})$ \\
      \Return $O$
  \end{algorithmic}
  \label{algo:kvs}
\end{algorithm}

\section{Details of Diffusion Distillation}
\label{sec:appendix_distill}

In \cref{sec:ablation}, we provide a brief ablation study of the proposed progressive flow distillation with a case study. Here, we present a more detailed comparison with additional test cases and also include ablations of Huber loss and Phase 1 fine-tuning.

\begin{figure*}[t]
  \centering
  \includegraphics[width=\linewidth]{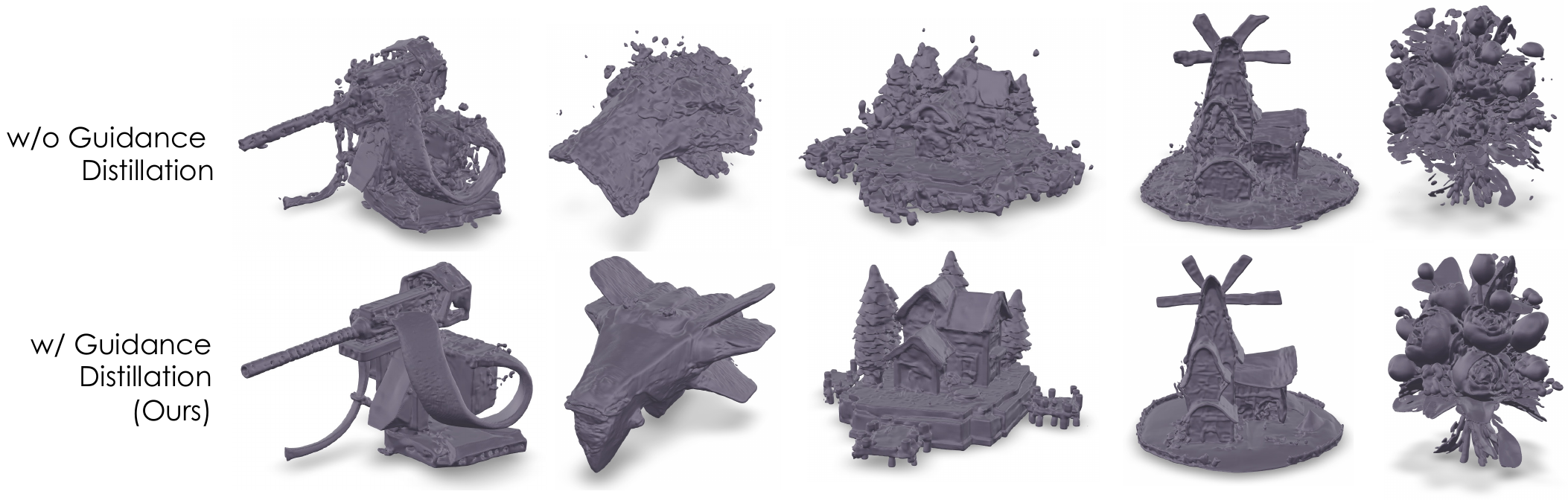}
   \caption{Visual comparison of models \textbf{with and without guidance distillation warmup}. The adversarial fine-tuning and Phase1 fine-tuning are not adopted. It demonstrates that the guidance distillation warmup is essential for successful distillation.}
   \label{fig:gd_ablation}
\end{figure*}

\textbf{Guidance Distillation Warmup.} \cref{fig:gd_ablation} shows the results without guidance distillation warmup. We observe significant degradation in results when guidance distillation is omitted, confirming the effectiveness of our strategy.

\begin{figure*}[t]
  \centering
  \includegraphics[width=\linewidth]{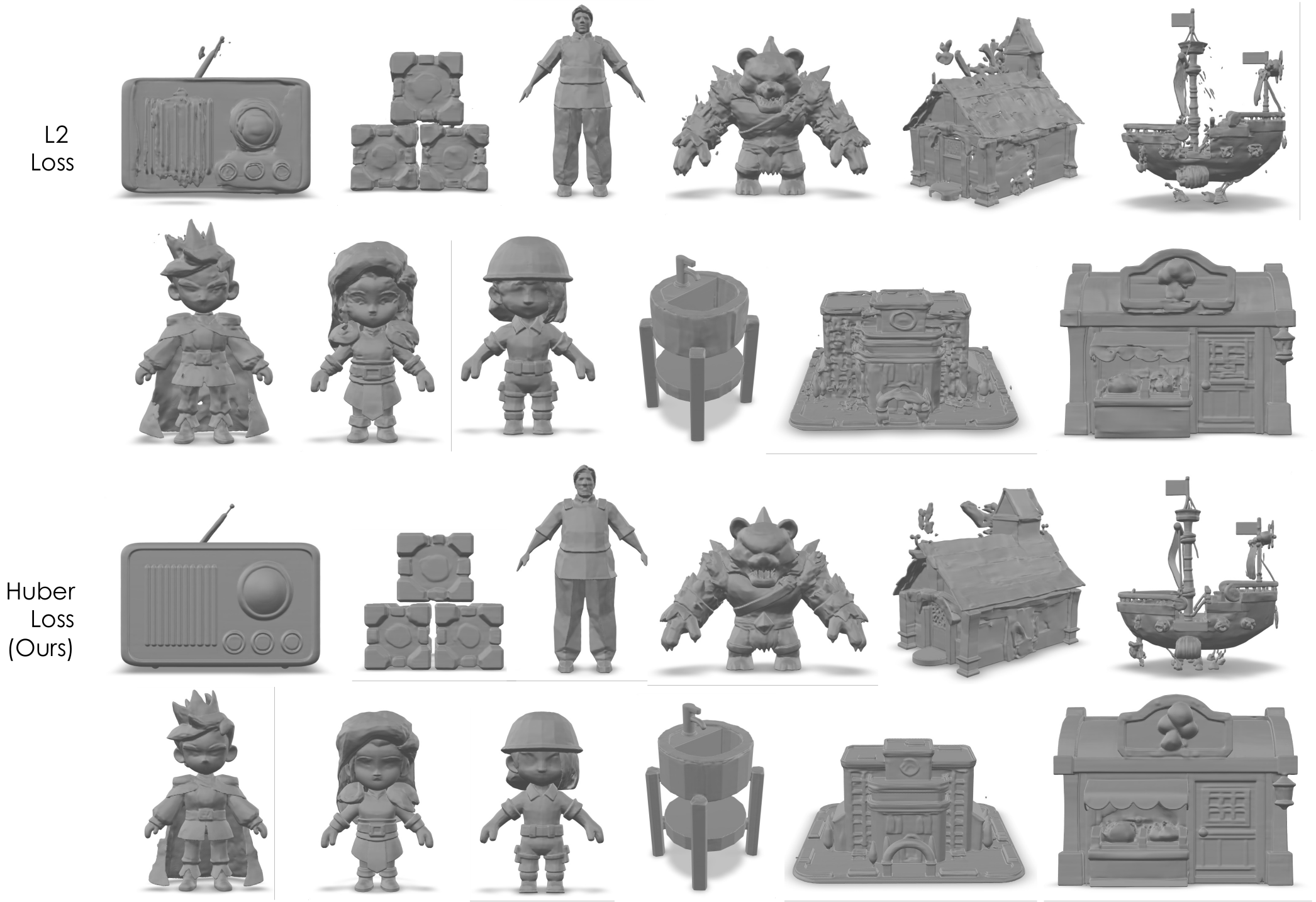}
   \caption{Visual comparison of models trained with \textbf{L2 and huber loss}. The adversarial fine-tuning and Phase1 fine-tuning are not adopted. It demonstrates that the huber loss is significantly better than l2 loss, which we hypothesis that is due to huber loss is less sensitive to outliers so that stablizes the training and makes results better.}
   \label{fig:huber_vs_l2}
\end{figure*}

\textbf{Huber Loss vs L2 Loss.} \cref{fig:huber_vs_l2} shows a visual comparison between models trained with L2 and Huber loss. While L2 loss generates reasonable results, the quality is noticeably inferior to that of the model trained with Huber loss. For example, certain structures, like the radio and several houses, are broken in the L2 model. We hypothesize that it is because  Huber loss is less sensitive to outliers, thus stabilizing the training and improving the results.

\begin{figure*}[t]
  \centering
  \includegraphics[width=\linewidth]{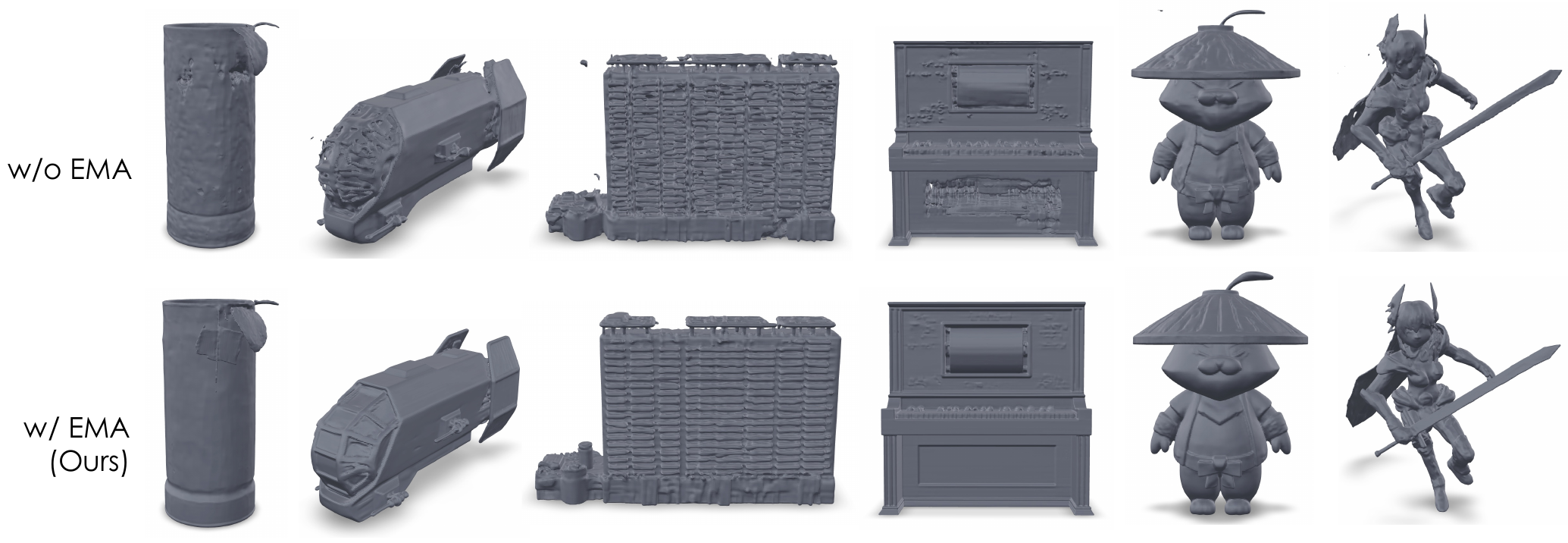}
   \caption{Visual comparison of models trained \textbf{with and without EMA}. The adversarial fine-tuning and Phase1 fine-tuning are not adopted. It demonstrates that the meshes tend to be broken without EMA.}
   \label{fig:ema_ablation}
   \foobar{5mm}
\end{figure*}

\textbf{EMA of Target Network.} \cref{fig:ema_ablation} compares models trained with and without EMA. Both models were fine-tuned from a guidance-distilled model using consistency flow distillation, with no Phase 1 or adversarial fine-tuning. It can be seen that the meshes are broken without EMA, highlighting the importance of EMA for stability.

\begin{figure*}[t]
  \centering
  \includegraphics[width=\linewidth]{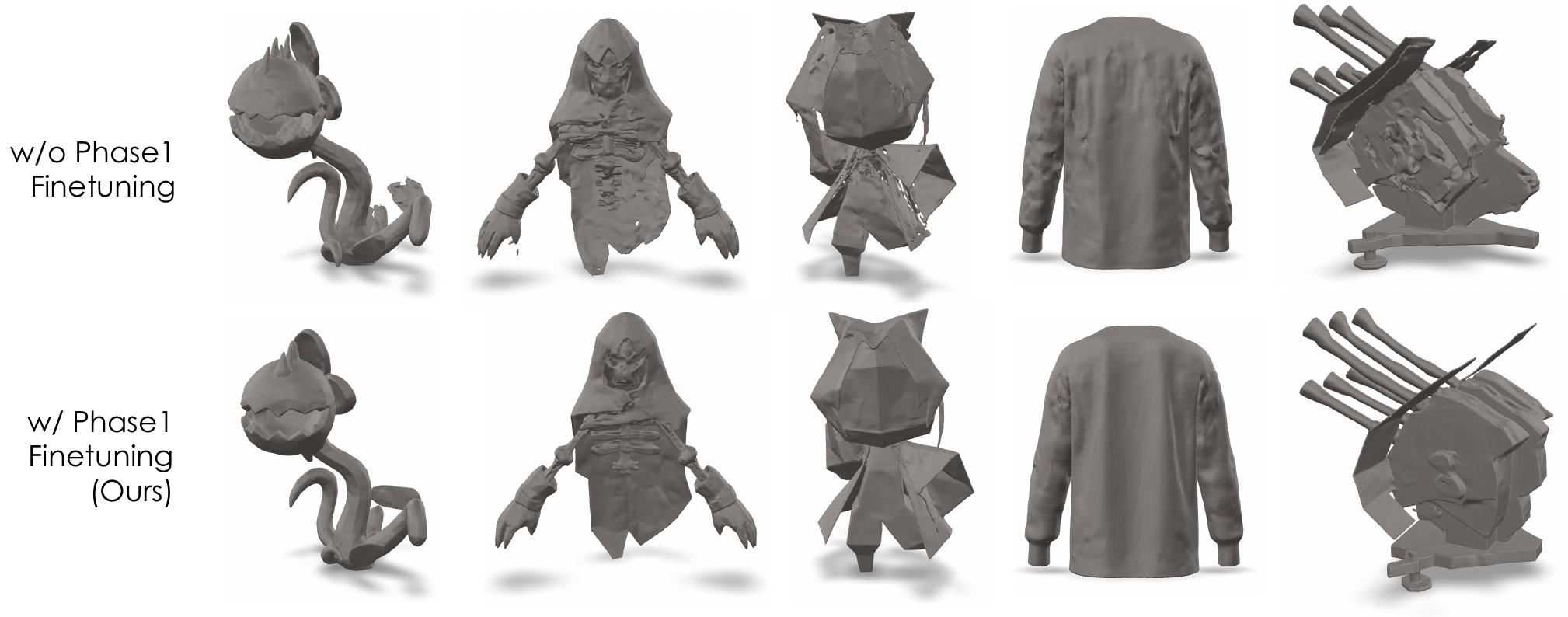}
   \caption{Visual comparison of models \textbf{with and without guidance distillation warmup}. The adversarial fine-tuning and Phase1 fine-tuning are not adopted. It demonstrates that the guidance distillation warmup is essential for successful distillation.}
   \label{fig:phase1_ablation}
\end{figure*}

\textbf{Phase 1 Fine-tuning.} During consistency flow distillation, we follow PCM~\cite{wang2025phased} to divide the total trajectory into 5 phases and force the model to predict different targets at each phase. However, there is a training-test gap as the model needs to predict final target during the inference. To address this, we propose Phase 1 fine-tuning after Phase 5 pretraining. We empirically find that this strategy slightly improves performance, as shown in \cref{fig:phase1_ablation}.

\begin{figure*}[t]
  \centering
  \includegraphics[width=\linewidth]{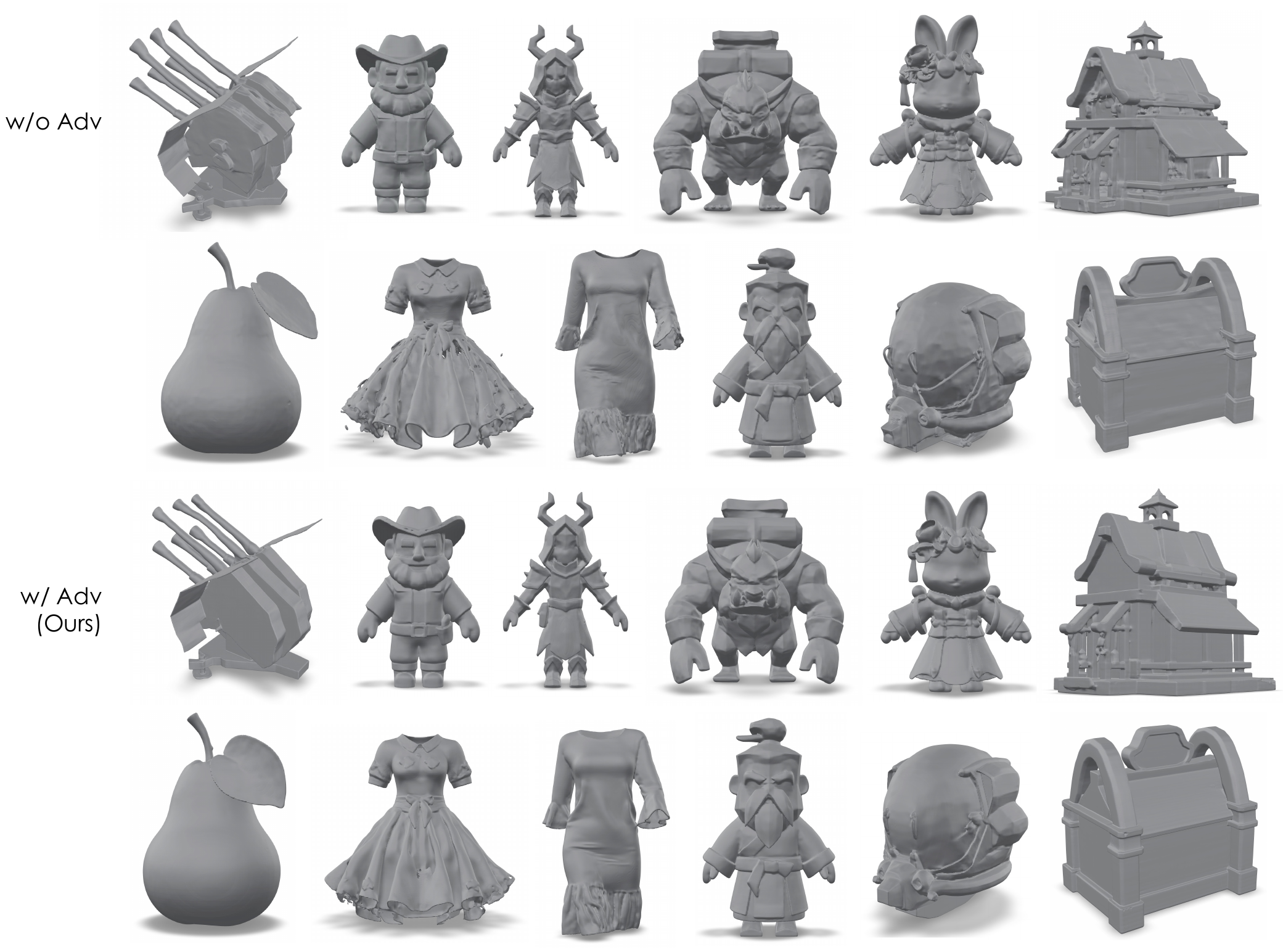}
   \caption{Visual comparison of models \textbf{with and without adversarial finetuning}. All other distillation stages are used. It demonstrates that the predicted meshes are more accurate and smooth after adversarial finetuning.}
   \label{fig:adv_results}
\end{figure*}

\textbf{Adversarial Fine-tuning.} The comparison between models with and without adversarial fine-tuning is shown in \cref{fig:adv_results}. It is evident that adversarial fine-tuning helps improve surface smoothness, corrects detail generation, and fixes mesh holes.

\begin{figure*}[t]
  \centering
  \includegraphics[width=0.9\linewidth]{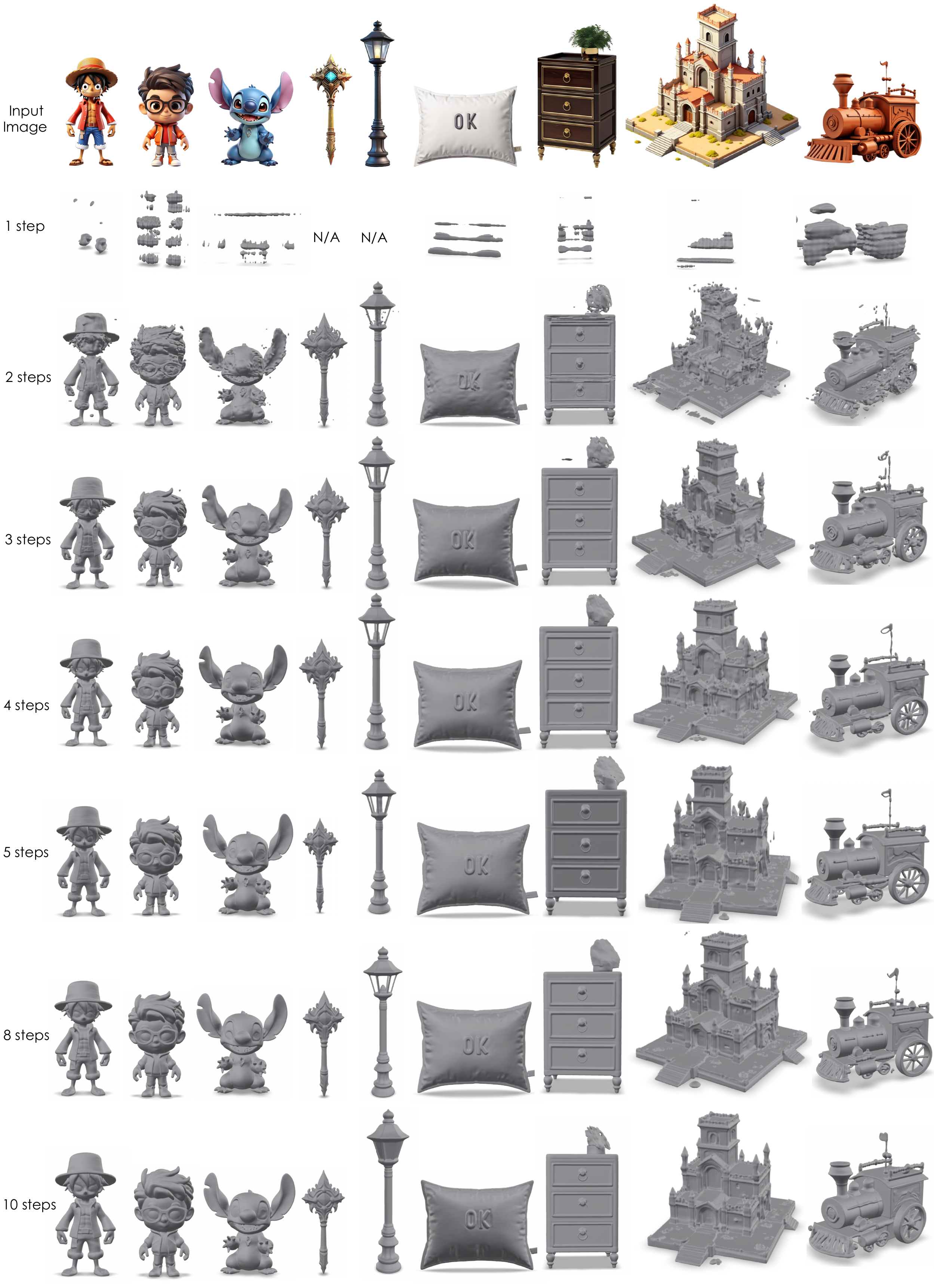}
   \caption{Visual comparison of \shortname generation results with different sampling steps. }
   \label{fig:multi_steps}
\end{figure*}

\textbf{Effect of Sampling Steps.} As shown in \cref{fig:multi_steps}, our method demonstrates the ability to generate rough results with just 2 steps, and simple objects can be effectively generated within 3 steps.

\section{More Results}

\begin{figure*}[t]
  \centering
  \includegraphics[width=\linewidth]{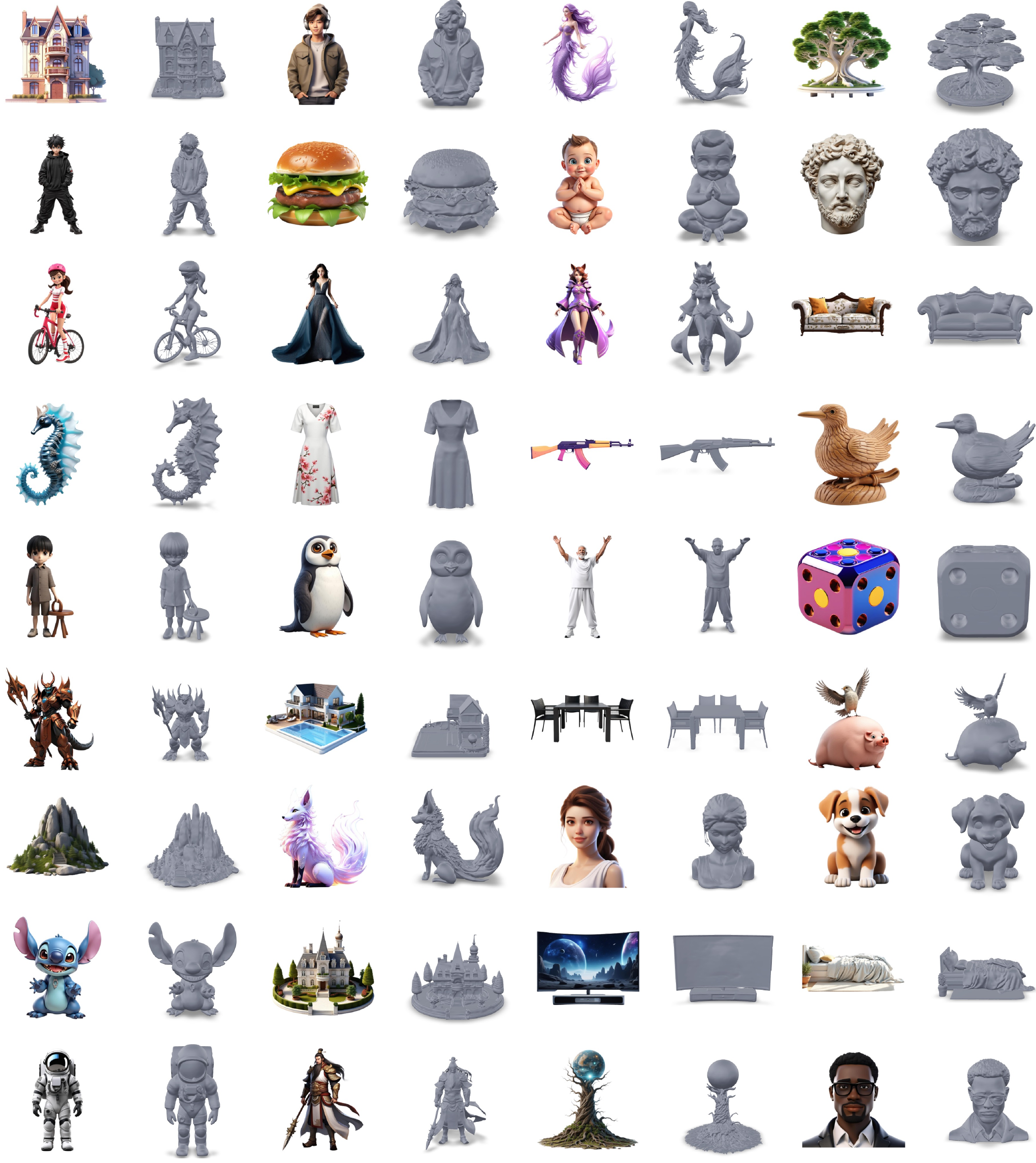}
   \caption{Shape generation results of \hyshape Turbo distilled with the proposed \shortname. Image prompts are generated by HunyuanDiT~\cite{li2024hunyuandit}. The number of inference steps is 5.}
   \label{fig:more_dit_results}
\end{figure*}

\textbf{Shape Generation Results.} \cref{fig:more_dit_results} presents a set of shape generation results from \hyshape Turbo, which has been distilled using the proposed \shortname framework. Our model achieves fast generation with only 5 diffusion sampling steps and ultra-fast decoding, while maintaining high-quality meshes across a variety of categories.

\begin{figure*}[t]
  \centering
  \includegraphics[width=\linewidth]{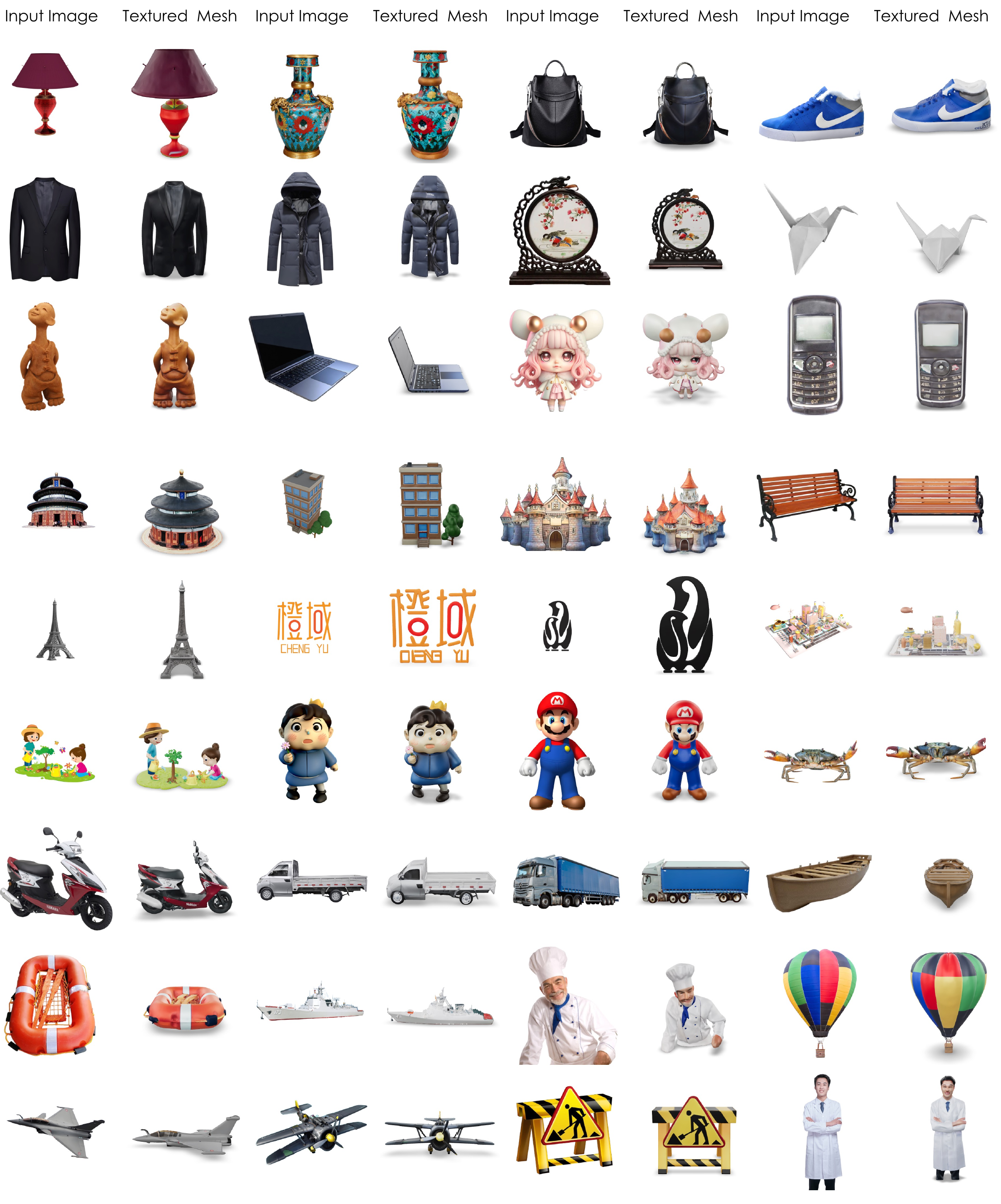}
   \caption{Texture generation results of \hyshape Turbo distilled with the proposed \shortname and Hunyuan3D-Paint-2~\cite{zhao2021large}. Image prompts are generated by HunyuanDiT~\cite{li2024hunyuandit}. The number of inference steps is 5. }
   \label{fig:more_dit_results_with_texture}
\end{figure*}

\textbf{Compatibility with Texture Generation.} \cref{fig:more_dit_results_with_texture} showcases texture generation results for meshes produced by \hyshape Turbo, distilled with \shortname. It is evident that the meshes generated by our method are fully compatible with texture generation, demonstrating its versatility.

\begin{figure*}[t]
  \centering
  \includegraphics[width=\linewidth]{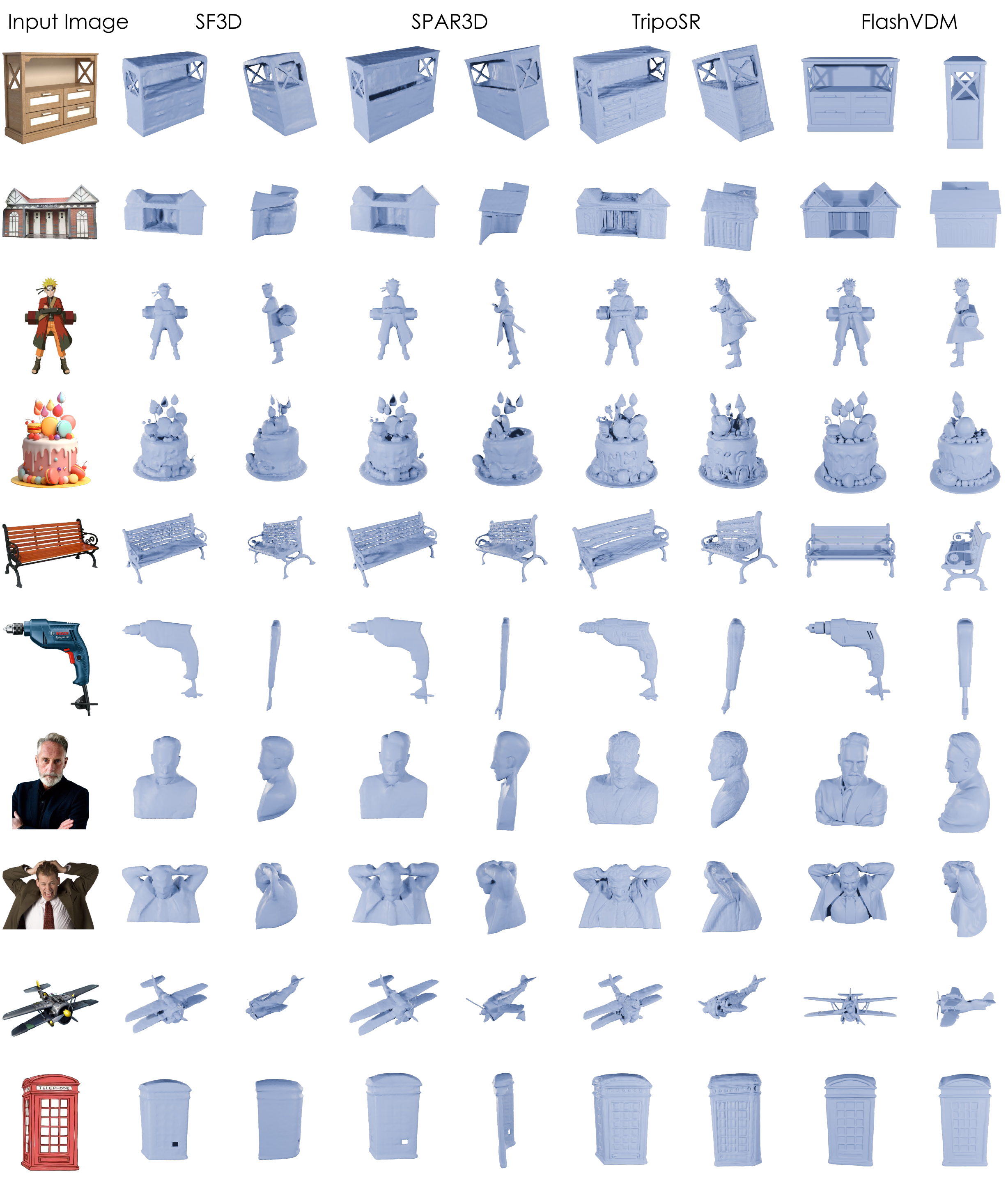}
   \caption{Comparison between \shortname (\hyshape Turbo) 5 steps and other fast 3D generation methods. }
   \label{fig:more_cmp_fast}
\end{figure*}

\textbf{Comparison with Other Methods.} \cref{fig:more_cmp_fast} compares \shortname with other fast 3D generation methods. The results highlight that our method consistently outperforms existing approaches across a broad range of input types.

\section{Limitations and Future Works.} 

In this work, we have significantly accelerated both VAE decoding and diffusion sampling. Despite these improvements, there are still areas that could be further enhanced. For instance, our PyTorch implementation contains several indexing operations, which can slow down the GPU pipeline. Operator fusion and more efficient memory access strategies could be promising directions for optimization. Additionally, exploration of locality of vecset would also be an interesting direction.
Regarding diffusion sampling, single-stage distillation may be preferable, as the current multi-stage approach is complex and introduces cascade errors, which limit its performance potential. Furthermore, while our investigation of adversarial finetuning shows promising results, further research could focus on continuously  utilizing real 3D data with adversarial finetuning or even reinforcement learning, a direction we believe holds significant promise. Lastly, as VAE inference time is reduced, the proportion of time spent on diffusion sampling increases. This suggests that exploring one-step distillation could be a valuable avenue for future research.

\clearpage